\documentclass[nohyperref]{article}

\usepackage[square,sort,comma,numbers]{natbib}
\usepackage[preprint]{neurips_2026}

\usepackage[utf8]{inputenc} % allow utf-8 input

\usepackage[T1]{fontenc}    % use 8-bit T1 fonts
\usepackage{hyperref} 
\usepackage{url}            % simple URL typesetting

\usepackage{booktabs}       % professional-quality tables
\usepackage{amsfonts}       % blackboard math symbols
\usepackage{nicefrac}       % compact symbols for 1/2, etc.
\usepackage{microtype}      % microtypography
\usepackage{xcolor}         % colors
\usepackage{amsmath}
\usepackage{amsthm}
\usepackage{xspace}
\usepackage{enumitem}
\usepackage[table]{xcolor}
\usepackage{pgfplots}

\usepackage{booktabs}
\usepackage{algorithm} \usepackage{algpseudocode}
\newcommand{\rmemsafe}{\textsc{RMemSafe}\xspace}
\pgfplotsset{compat=1.18}
\usepgfplotslibrary{groupplots}
\usepackage{amsmath, amssymb}
\usepackage{xcolor}
\usepackage{pgfplots}
\pgfplotsset{compat=1.18}
\usetikzlibrary{arrows.meta, patterns}
\usepackage{multirow}
\usepackage{pifont}
\newcommand{\cmark}{\ding{51}}
\newcommand{\xmark}{\ding{55}}
\newtheorem{proposition}{Proposition}
\newif\ifcommentsoff
\makeatletter
\newcommand{\comments}[3]{%
  \ifcommentsoff
    \@bsphack\@esphack% Handles spacing invisibly
  \else
    \@bsphack
    \textcolor{#2}{[#1: #3]}%
    \@esphack
  \fi
}
\makeatother

\title{Reliability-Gated Source Anchoring for Continual Test-Time Adaptation}

\author{%
  \textbf{Vikash Singh\textsuperscript{1}} \quad
  \textbf{Debargha Ganguly\textsuperscript{1}} \quad
  \textbf{Weicong Chen\textsuperscript{1}} \quad
  \textbf{Sabyasachi Sahoo\textsuperscript{2,3}} \\
  \textbf{Sreehari Sankar\textsuperscript{1}} \quad
  \textbf{Biyao Zhang\textsuperscript{1}} \quad
  \textbf{Mohsen Hariri\textsuperscript{1}} \quad
  \textbf{Shouren Wang\textsuperscript{1}} \\
  \textbf{Osama Zafar\textsuperscript{1}} \quad
   \textbf{Christian Gagn\'e\textsuperscript{2,3}} \quad
  \textbf{Vipin Chaudhary\textsuperscript{1}} \\[0.3em]
  \textsuperscript{1}Case Western Reserve University \quad
  \textsuperscript{2}Universit\'e Laval \quad
  \textsuperscript{3}Mila - Qu\'ebec AI Institute \\[0.3em]
  \texttt{vikash@case.edu}
}
\begin{document}

\maketitle

\begin{abstract}

Continual test-time adaptation (CTTA) updates a pretrained model online on an unlabeled, non-stationary stream while anchoring it to a frozen source checkpoint. This anchor is useful only when the source remains reliable. On CCC-Hard, however, a ResNet-50 source falls to approximately $1.3\%$ top-$1$ accuracy, while existing source-anchored CTTA methods continue applying the same anchor strength. We call this failure mode \emph{blind anchoring} and propose \rmemsafe, a reliability-gated extension of ROID that uses the frozen source's normalized predictive entropy to attenuate all explicit source-coupled uses in the objective. When the source posterior approaches uniformity, the gate closes: the source anchor and agreement filter vanish, and the objective reduces to a source-agnostic fallback comprising ROID's base losses plus marginal calibration. Combined with ASR, \rmemsafe achieves the lowest error on $8$ of $9$ matched-split continual-corruption cells and is the best reset-based method on all $9$, improving ROID{+}ASR by $1.05$~pp on ResNet-50 and $0.48$~pp on ViT-B/16. A controlled source-degradation sweep shows a $1.13{\times}$ shallower harm slope than ROID{+}ASR, consistent with the graceful-decay prediction. The entropy gate detects high-entropy source collapse, not confidently wrong low-entropy sources; this scope is explicitly evaluated and discussed.

\end{abstract}

\section{Introduction}
\label{sec:intro}

A model deployed against a shifting, unlabeled data stream faces a basic safety question: can the adaptation procedure be prevented from driving the model into states worse than those it started in? Continual test-time adaptation (CTTA) addresses this by updating a pre-trained model online while anchoring it to its frozen source~\cite{wang2022cotta,niu2022efficient,niu2023towards,marsden2024universal,lim2026and,prabhu2020gdumb}, the anchor preventing runaway drift under noisy pseudo-labels. The anchor carries a silent, runtime-unchecked precondition: that the source remains a meaningful reference throughout adaptation. On the hardest level of the CCC benchmark~\cite{press2023rdumb,lim2026and}, the ResNet-50 source's top-1 accuracy is $\sim\!1\%$; yet every prior CTTA method we evaluate continues to pull the adapting model toward this degraded reference with an unconditional $\ell_2$ penalty of fixed strength. This blind anchoring is a systematic failure mode of current CTTA methods, and fixed-strength trust in the source can be replaced with a lightweight runtime check that recovers an analytical graceful-decay property.

\providecommand{\phdefinecolors}{%
  \definecolor{phAcc}{HTML}{1F4E79}%    dark blue (source accuracy)
  \definecolor{phPrior}{HTML}{B71C1C}%  red (prior methods: flat anchor)
  \definecolor{phOurs}{HTML}{2E7D32}%   green (ours: scaled anchor)
  \definecolor{phDanger}{HTML}{FFCDD2}% pale red (catastrophic regime)
}
\phdefinecolors

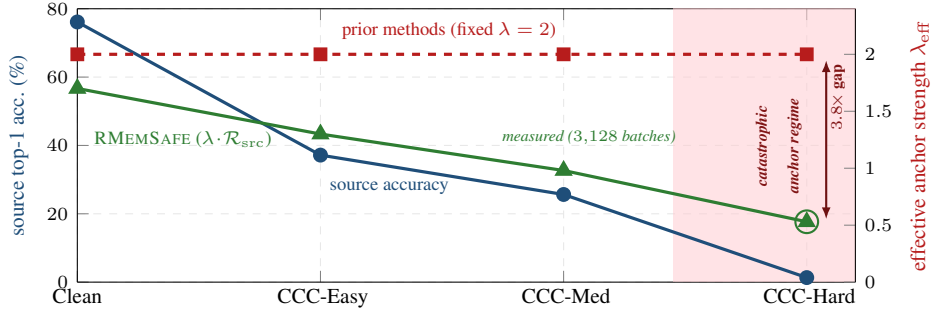
\begin{figure}[htbp]
  \centering
  \begin{tikzpicture}
    \begin{axis}[
      name=ax,
      width=0.85\linewidth, height=5.2cm,
      xmin=0, xmax=3.2,
      ymin=0, ymax=80,
      xtick={0, 1, 2, 3},
      xticklabels={Clean,\ CCC-Easy,\ CCC-Med,\ CCC-Hard},
      x tick label style={font=\footnotesize, yshift=1pt},
      ylabel={\textcolor{phAcc}{source top-1 acc.\ (\%)}},
      ylabel style={font=\footnotesize},
      ytick={0,20,40,60,80},
      tick label style={font=\scriptsize},
      axis y line*=left,
      grid=major, grid style={dashed, gray!18},
      clip=false,
    ]
      % Shade the "catastrophic anchoring" regime on the right
      \fill[phDanger, opacity=0.55]
        (axis cs:2.45,0) rectangle (axis cs:3.2,80);
      % Two-line vertical label, carefully sized to fit inside the
      % shaded strip without touching the top axis or the dashed line.
      \node[font=\tiny\itshape\bfseries, text=phPrior!70!black,
            rotate=90, anchor=center, align=center]
        at (axis cs:2.82, 40)
        {catastrophic};
      \node[font=\tiny\itshape\bfseries, text=phPrior!70!black,
            rotate=90, anchor=center, align=center]
        at (axis cs:2.95, 40)
        {anchor regime};

      % Source top-1 accuracy (from Table 1, ResNet-50)
      \addplot[phAcc, very thick, mark=*, mark size=2.2pt]
        coordinates {
          (0, 76.13)   % clean ImageNet val
          (1, 37.17)   % CCC-Easy: 100 - 62.83
          (2, 25.65)   % CCC-Med:  100 - 74.35
          (3, 1.33)    % CCC-Hard: 100 - 98.67
        };

      % Direct label on the source-accuracy curve (no legend box).
      \node[font=\scriptsize, text=phAcc, anchor=south west]
        at (axis cs:1.0, 23.0) {source accuracy};
    \end{axis}

    % Right y-axis: effective anchor strength
    \begin{axis}[
      at={(ax.south west)}, anchor=south west,
      width=0.85\linewidth, height=5.2cm,
      xmin=0, xmax=3.2,
      ymin=0, ymax=2.4,
      axis y line*=right,
      axis x line=none,
      ylabel={\textcolor{phPrior}{effective anchor strength $\lambda_{\mathrm{eff}}$}},
      ylabel style={font=\footnotesize},
      ytick={0, 0.5, 1.0, 1.5, 2.0},
      tick label style={font=\scriptsize},
      clip=false,
    ]
      % Prior methods: flat anchor (ROID+ASR, RDumb, ETA/EATA+ASR)
      \addplot[phPrior, very thick, dashed, mark=square*,
               mark size=2.0pt, mark options={solid, fill=phPrior}]
        coordinates {
          (0, 2.0) (1, 2.0) (2, 2.0) (3, 2.0)
        };

      % Ours: R_src-gated. Values: R_src*lambda.
      % R_src(clean)~0.85, R_src(Easy)~0.65, R_src(Med)~0.49,
      % R_src(Hard)=0.263 measured.
      \addplot[phOurs, very thick, mark=triangle*,
               mark size=2.8pt, mark options={fill=phOurs}]
        coordinates {
          (0, 1.70)   % 2.0 * 0.85
          (1, 1.30)   % 2.0 * 0.65
          (2, 0.98)   % 2.0 * 0.49
          (3, 0.53)   % 2.0 * 0.263   (measured)
        };

      % Direct labels on the two anchor curves (clear space above each).
      \node[font=\scriptsize, text=phPrior, anchor=south west]
        at (axis cs:1.05, 2.05)
        {prior methods (fixed $\lambda=2$)};
      \node[font=\scriptsize, text=phOurs, anchor=south west]
        at (axis cs:0.03, 1.1)
        {\rmemsafe ($\lambda\!\cdot\!\mathcal{R}_{\mathrm{src}}$)};

      % Highlight the measured CCC-Hard point (small ring around the marker).
      \node[circle, draw=phOurs, line width=0.8pt, inner sep=1pt,
            minimum size=3mm] (mpt) at (axis cs:3, 0.53) {};
      % "measured" tag: place BELOW the point so it sits in the empty
      % strip between the green curve and the axis, never touching the
      % source-accuracy curve.
      \node[font=\tiny\itshape, text=phOurs, anchor=north east]
        at (axis cs:2.5, 1.42)
        {measured ($3{,}128$ batches)};

      % The "gap" that RMemSafe opens up, drawn in the clear strip
      % just inside the right edge.
      \draw[<->, >={Latex[length=1.6mm]}, thick, phPrior!60!black]
        (axis cs:3.08, 0.55) -- (axis cs:3.08, 1.95);
      \node[font=\tiny\bfseries, text=phPrior!70!black,
            anchor=west, rotate=90]
        at (axis cs:3.13, 1.25) {$3.8\!\times$ gap};
    \end{axis}

  \end{tikzpicture}
\caption{\textbf{Continual test-time adaptation under collapsing source reliability.} CTTA anchors the adapter to its frozen source, presuming the source stays a meaningful reference. \textcolor{phAcc}{Blue:} on CCC, frozen RN-50 source top-$1$ collapses from $76\%$ (clean ImageNet) to $1.3\%$ (CCC-Hard). \textcolor{phPrior}{Red dashed:} prior reset-based methods (ROID/ETA/EATA{+}ASR, ROID{+}RDumb) hold $\lambda{=}2$ across all severities, pulling the adapter toward near-noise output. \textcolor{phOurs}{Green:} \rmemsafe multiplies $\lambda$ by the runtime reliability $\mathcal{R}_{\mathrm{src}}{=}\max(0,1{-}\mathcal{H}_{\mathrm{src}})$ from the source's own posterior entropy. The CCC-Hard $\lambda_{\mathrm{eff}}{=}0.53$ is measured ($3{,}128$-batch trace, App.~Fig.~\ref{fig:trace}); the others follow the monotone entropy-accuracy relation. In the catastrophic regime (shaded), \rmemsafe attenuates the anchor $3.8\times$; this attenuation is the analytical safety property demonstrated under controlled degradation (\S\ref{sec:experiments}, App.~\ref{app:harm_slope}). Matched-split gains on standard CCC reflect the integrated objective; the gate's safety role is isolated on the controlled source-degradation axis (App.~\ref{app:cumulative-ablation}).}
\label{fig:phenomenon}
\vspace{-4ex}
\end{figure}
\rmemsafe instantiates this check. It gates all explicit source-coupled uses in the ROID~\cite{marsden2024universal} optimization: the $\ell_2$ anchor toward $\theta^{*}_{\mathrm{src}}$, the cosine-based source-expert agreement filter, and the source-divergence factor inside the anchor. All three are multiplicatively modulated by a single source-reliability scalar $\mathcal{R}_{\mathrm{src}} = \max(0, 1 - \mathcal{H}_{\mathrm{src}})$ derived from source entropy. Because gating alone is not sufficient (the remaining loss must not itself collapse when the gate closes), the gate is composed of three auxiliary stabilizers adapted from standard practice: marginal calibration, a confidence-scaled learning rate, and a decoupled confidence-interpolated flip at inference. The resulting method admits an analytical graceful-decay property (Proposition~\ref{prop:graceful}): when source entropy saturates, the source-dependent terms vanish, and the total objective reduces to the base CTTA loss plus the marginal-calibration term, independent of the frozen source parameters. On CCC-Hard, the gate attenuates the runtime anchor by a factor of $3.8\!\times$ relative to prior methods with a fixed $\lambda=2$. This attenuation is the method's analytical safety property; its empirical signature is a shallower harm slope under controlled source degradation, while the matched-split benchmark improvements reflect the integrated objective operating jointly.

Two evaluation axes confirm the property. On matched-split evaluations across nine continual-corruption benchmark cells (two architectures, three CCC difficulty levels, two CIN-C orderings, and IN-C 20-revisit), \rmemsafe achieves the lowest error on 8 of 9 cells, reducing mean CCC error over ROID{+}ASR, the strongest matched-split prior baseline, by $1.05$~pp on ResNet-50 and $0.48$~pp on ViT-B/16 (paired $t$, pooled $p{<}10^{-16}$, $95\%$ CI $[-1.16,-0.94]$, Cohen's $d_{z}{=}{-}3.65$; App.~\ref{app:significance}, Table~\ref{tab:effect_size}). In a controlled source-degradation experiment that varies clean-test source accuracy from $S{=}0.75$ down to $S{=}0.12$ via Gaussian weight noise, \rmemsafe degrades more gracefully than ROID{+}ASR across every tested severity, with a harm slope of $11.43$~pp per unit of $S$ versus $12.92$~pp for ROID{+}ASR, a $1.13\!\times$ ratio in the direction predicted by Proposition~\ref{prop:graceful}.

An explicit scope is stated upfront. The reliability gate uses source \emph{entropy} as its sole reliability signal. This handles the dominant CTTA failure mode observed in practice, high-entropy source collapse as exemplified by CCC-Hard, but not \emph{confidently miscalibrated} sources: a source that is wrong without being uncertain would be deemed reliable by $\mathcal{R}_{\mathrm{src}}$. Designing a correctness-aware reliability signal is a direction for future work (\S\ref{sec:discussion}). We flag this scope here so the method's claims are understood narrowly. We confirm this empirically: under a permuted-class low-entropy wrong source \rmemsafe loses its advantage as predicted, while in the high-entropy regime it preserves it (Table~\ref{tab:wrong_source}). \rmemsafe is graceful-decay \emph{with respect to entropy-detectable source failure}.

This paper advances graceful-decay CTTA through the following key contributions:
\begin{itemize}[nosep,leftmargin=1.2em]
  \item \textbf{Diagnosis of blind anchoring.} On CCC-Hard, prior methods anchor at full strength to a source with ${\sim}1\%$ top-$1$ accuracy: a systematic CTTA failure mode we name and characterize (Figure~\ref{fig:phenomenon}).
\item \textbf{Reliability-gated CTTA with a graceful-decay guarantee.} To our knowledge, \rmemsafe is the first CTTA method to gate all explicit source-coupled uses by a runtime reliability signal derived from the frozen source and to admit an analytical graceful-decay property (Prop.~\ref{prop:graceful}); the integrated objective keeps the fallback regime well-defined.
  % ; a harm-slope
  % decomposition predicts the observed $1.13{\times}$ ratio
  % (App.~\ref{app:harm_slope}).
  
  \item \textbf{Empirical confirmation at real and controlled degradation.} $8/9$ wins on matched-split benchmarks (per-cell paired $t$, CIs, $d_z$ in App.~\ref{app:significance}); shallower harm slope than ROID{+}ASR under a controlled source-quality sweep, within an explicitly stress-tested scope (App.~\ref{app:wrong_source}).
  \item \textbf{Diagnosis of a reset-paradigm failure mode.} On CCC-Hard with ViT-B/16, every reset-based method underperforms non-reset ROID, characterized at matched-split granularity for the first time (\S\ref{sec:exp:reset-paradigm}); a reliability-gated $\tau$-sweep partially closes the gap (App.~\ref{app:gate_tau}).
\end{itemize}

\section{Related Work}
\label{sec:related_work}

\textbf{Test-Time Adaptation.} TTA~\cite{sun2020ttt} adapts pre-trained models from unlabeled streams via test-time batch-norm recalibration~\cite{schneider2020improving}, entropy minimization~\cite{wang2021tent}, Fisher-regularized updates~\cite{niu2022efficient}, self-supervised auxiliary updates~\cite{liu2021tttpp}, prototype-based classifier adjustment~\cite{iwasawa2021t3a} and per-instance test-time augmentation~\cite{zhang2022memo}; non-saturating surrogates, such as the soft likelihood ratio~\cite{mummadi2021slr} inherited via ROID, replace entropy where it saturates. DeYO~\cite{lee2024deyo} shows entropy-based selection fails under disentangled spurious factors, and COME~\cite{zhang2025come} addresses entropy's overconfidence pathology via Dirichlet-prior conservative entropy. We use entropy as a frozen-source reliability proxy rather than as a self-training target for the adapting expert; this changes the failure mode, but does not eliminate low-entropy miscalibration, which we evaluate in \S\ref{sec:discussion:scope} and Appendix~\ref{app:wrong_source}. Standalone TTA still suffers catastrophic forgetting and error accumulation under continuous, non-stationary shifts~\cite{liang2023ttasurvey}.

\textbf{Continual Test-Time Adaptation.} CTTA stabilizes continuous adaptation through stochastic restoration~\cite{wang2022cotta}, sample filtering~\cite{niu2023towards}, instance-aware normalization under temporal correlation~\cite{gong2022note,yuan2023rotta}, symmetric-cross-entropy mean-teacher training~\cite{dobler2023rmt} (whose SymCE term we inherit in our source-agnostic fallback), momentum ensembling~\cite{marsden2024universal}, Kalman filtering~\cite{lee2024continual}, memory-efficient self-distilled regularization~\cite{song2023ecotta}, adaptive collapse-sensing~\cite{hoang2024petta}, and robust pseudo-labeling self-training~\cite{rusak2022rpl,tzeng2017adda}. A critical limitation along this line is the implicit assumption that the static source remains a reliable anchor, which fails when severe corruption catastrophically confuses the source. \rmemsafe addresses this ``blind anchoring'' flaw by dynamically gating source agreement and anchor strength based on localized source reliability, with an analytically derived graceful-decay property under source collapse.

\textbf{Long-Term Adaptation and Resets.} Long-term severe shifts (e.g., CCC-Hard~\cite{press2023rdumb,mishra2026rdumbdriftawarecontinualtesttime}) cause CTTA models to collapse, motivating the reset paradigm: periodic parameter restoration (RDumb~\cite{press2023rdumb}) or adaptive resets (ASR~\cite{lim2026and}). Resets arrest collapse but may erase acquired target knowledge, and a reset toward a catastrophically confused source can itself be harmful (\S\ref{sec:exp:reset-paradigm}). \rmemsafe is complementary: it attenuates unreliable source-guided updates before they accumulate, improving adaptation-phase stability, and combines effectively with ASR. The CCC and ImageNet-C benchmarks~\cite{press2023rdumb,hendrycks2019benchmarking} provide the corruption streams used in our evaluation.

\section{Method: Reliability-Gated Test-Time Adaptation}
\label{sec:method}
% when the frozen source's posterior becomes uniform,
% every source-dependent loss term should multiplicatively decay to
% zero, leaving an objective whose gradient no longer references the
% collapsed source. Proposition~\ref{prop:graceful} formalises this;
% the rest of this section constructs the gate and the stabilisers
% that keep the fallback regime well-behaved.
When the source posterior is uniform, every source-dependent term should vanish multiplicatively, thereby dropping the source from the gradient (Prop.~\ref{prop:graceful}); we now construct the gate and stabilizers that keep this fallback well-behaved.

\begin{figure}[ht]
    \centering
    \includegraphics[width=\linewidth]{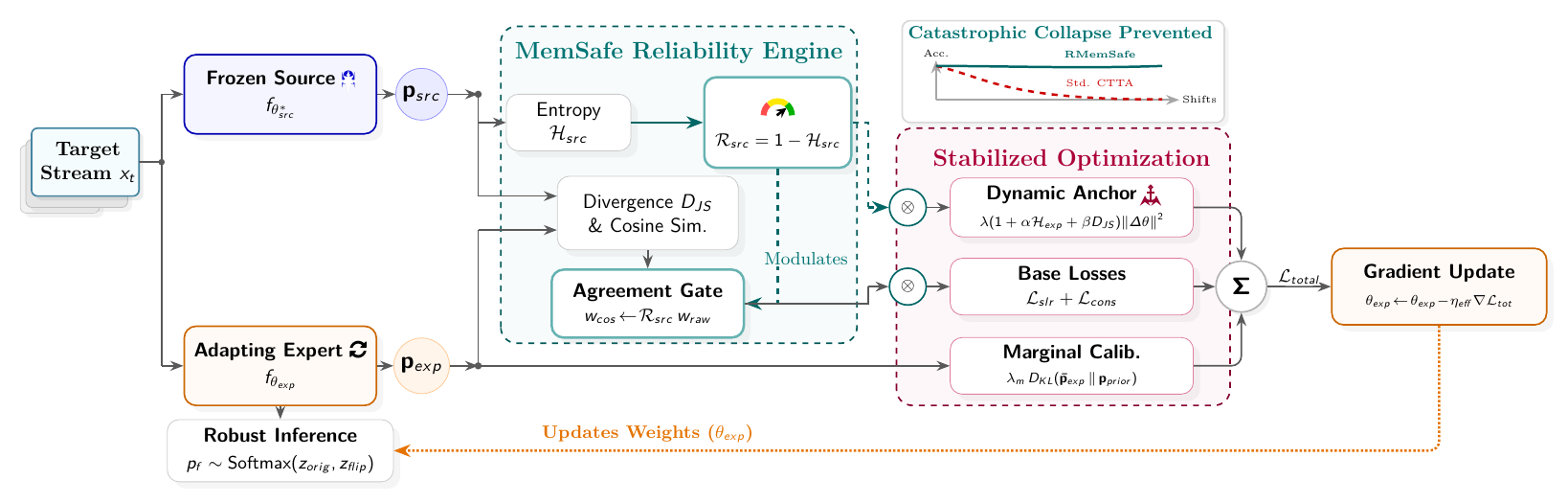}
    \caption{\textbf{Overview of \rmemsafe.} The reliability engine derives the source-reliability gate $\mathcal{R}_{src}=1-\mathcal{H}_{src}$ from the frozen source's entropy. $\mathcal{R}_{src}$ gates all explicit source-coupled uses: the Dynamic Anchor, the agreement filter (interpolating between source agreement and pass-through), and the source-divergence scaling inside the anchor, while leaving the base ROID losses and marginal calibration ungated. At inference, a confidence-interpolated flip average produces the final prediction. When the source collapses, $\mathcal{R}_{src}\!\rightarrow\!0$ and the total loss reduces to a non-collapsing source-agnostic fallback regime (Proposition~\ref{prop:graceful}).}
    \label{fig:rmemsafe-overview}
\end{figure}

\label{sec:method:setup}
Let $f_{\theta^{*}_{\mathrm{src}}}$ denote a pre-trained source model with frozen parameters $\theta^{*}_{\mathrm{src}}$ trained on a source distribution $\mathcal{D}_S$. During continual test-time adaptation, the model observes an unlabeled stream $x_t \in \mathcal{X}_T$ drawn from a shifted, non-stationary target distribution $\mathcal{D}_T$ and continuously updates expert parameters $\theta_{\mathrm{exp}}$ (initialized as $\theta_{\mathrm{exp}} \gets \theta^{*}_{\mathrm{src}}$) to minimize empirical risk on the stream. We build on the ROID framework~\cite{marsden2024universal}, which optimizes a soft-likelihood-ratio loss $\mathcal{L}_{\mathrm{slr}}$ and a consistency loss $\mathcal{L}_{\mathrm{cons}}$ on augmented views under diversity and certainty weighting, and additionally anchors the adapting parameters to the frozen source via a fixed-strength $\ell_2$ penalty. The design choice that motivates our method is that this anchor strength is \emph{unconditional} on source quality, if the frozen source's predictions collapse under distribution shift (on CCC-Hard, source top-1 accuracy is approximately $1\%$), the anchor continues to pull the adapting model toward a useless reference with full strength.

\textbf{The Reliability Gate.}
\label{sec:method:gate}
For each batch $x_t$ we compute the source and expert predictive distributions $\mathbf{p}_{\mathrm{src}} = f_{\theta^{*}_{\mathrm{src}}}(x_t)$ and $\mathbf{p}_{\mathrm{exp}} = f_{\theta_{\mathrm{exp}}}(x_t)$, and the normalized source entropy
\begin{equation}
\mathcal{H}_{\mathrm{src}} = -\tfrac{1}{\log C}\sum_{c=1}^C
p_{\mathrm{src}}^{(c)}\log p_{\mathrm{src}}^{(c)} \;\in\; [0,1],
\end{equation}
where $C$ is the number of classes. The \emph{source reliability gate} is $\mathcal{R}_{\mathrm{src}}= \max\!\bigl(0,\;1 - \mathcal{H}_{\mathrm{src}}\bigr).
\label{eq:rsrc}$
$\mathcal{R}_{\mathrm{src}}$ takes value $1$ when the source is maximally confident and $0$ when its posterior is uniform. \rmemsafe uses $\mathcal{R}_{\mathrm{src}}$ as a single scalar signal to modulate every source-dependent optimization term.

\textbf{Source-coupled uses gated by reliability.} $\mathcal{R}_{\mathrm{src}}$ is applied to all explicit places where ROID's optimization uses source information: the anchor, the agreement filter, and the source-divergence factor inside the anchor. \textbf{(i) Divergence-aware dynamic anchor.} Replacing the fixed $\ell_2$ penalty of ROID, we define
\begin{equation}
\mathcal{L}_{\mathrm{anch}} \;=\; \lambda\cdot\mathcal{R}_{\mathrm{src}}
\bigl(1 + \alpha\,\mathcal{H}_{\mathrm{exp}} + \beta\,D_{\mathrm{JS}}(\mathbf{p}_{\mathrm{src}}\|\mathbf{p}_{\mathrm{exp}})\bigr)\,
\|\theta_{\mathrm{exp}} - \theta^{*}_{\mathrm{src}}\|_2^{2},
\label{eq:anchor}
\end{equation}
where $\mathcal{H}_{\mathrm{exp}}$ is the expert's normalized entropy, $D_{\mathrm{JS}}$ is Jensen--Shannon divergence, and $\lambda,\alpha,\beta$ are fixed scalars. The outer $\mathcal{R}_{\mathrm{src}}$ prefactor makes the anchor pull toward $\theta^{*}_{\mathrm{src}}$ only when the source is reliable; the inner bracket makes the pull stronger when the expert is uncertain or diverges from the source. \textbf{(ii) Source-expert agreement gating.} The ROID base losses are gated by a cosine-agreement weight $w_{\mathrm{raw}} = w_{\min} + (1-w_{\min})\max(0, \cos(\mathbf{p}_{\mathrm{src}},\mathbf{p}_{\mathrm{exp}}))$\footnote{$\max(\cdot,0)$ is redundant since $p_{\mathrm{src}},p_{\mathrm{exp}}$ are softmax outputs and so $\cos\!\geq\!0$; we keep it as a numerical safeguard.}, and the gate itself is gated by $\mathcal{R}_{\mathrm{src}}$:
\begin{equation}
w_{\mathrm{cos}}= \mathcal{R}_{\mathrm{src}}\,w_{\mathrm{raw}} + (1-\mathcal{R}_{\mathrm{src}})\cdot 1,
\label{eq:wcos}
\end{equation}
which smoothly interpolates between source-expert agreement filtering (when $\mathcal{R}_{\mathrm{src}}\!\to\!1$) and unfiltered adaptation (when $\mathcal{R}_{\mathrm{src}}\!\to\!0$). Under an unreliable source, the agreement signal itself is untrustworthy, so the filter is disabled. \textbf{(iii) Divergence-scaled anchor strength.} The $\beta D_{\mathrm{JS}}$ term in \eqref{eq:anchor} is itself source-dependent: a large JS divergence between source and expert carries no information when the source is uninformative. It is gated by the same $\mathcal{R}_{\mathrm{src}}$ prefactor via its location inside $\mathcal{L}_{\mathrm{anch}}$, so its contribution decays to zero with source reliability.

\textbf{Auxiliary Stabilizers for the Fallback Regime.}
\label{sec:method:stabilizers}
Gating the source-dependent terms is necessary but not sufficient: when $\mathcal{R}_{\mathrm{src}}\!\to\!0$ the gated terms vanish, and the remaining loss must not itself collapse. We therefore include three auxiliary stabilizers, each adapted from standard practice in test-time adaptation. These are not claimed as contributions of this work; we describe them so that the fallback regime is fully specified.

\textbf{Marginal calibration.} To prevent class collapse without enforcing uniform batch marginals, we penalize the KL divergence between the current batch's class marginal $\bar{\mathbf{p}}_{\mathrm{exp}} = \tfrac{1}{B}\sum_b \mathbf{p}_{\mathrm{exp},b}$ and an EMA of historical class priors $\mathbf{p}_{\mathrm{prior}}$: $\mathcal{L}_{\mathrm{marg}}= D_{\mathrm{KL}}(\bar{\mathbf{p}}_{\mathrm{exp}}\,\|\,\mathbf{p}_{\mathrm{prior}}).$ This term is not source-dependent and is active regardless of $\mathcal{R}_{\mathrm{src}}$.

\textbf{Confidence-scaled learning rate.} To dampen destructive updates under high expert uncertainty, the effective learning rate is scaled as $\eta_{\mathrm{eff}} = \eta\cdot\bigl(\eta_{\min} + (1-\eta_{\min})(1-\mathcal{H}_{\mathrm{exp}})\bigr)$. This depends on expert uncertainty, not source reliability.

\textbf{Decoupled flip interpolation at inference.} For a test input $x_t$ and its horizontal flip $\tilde{x}_t$, the final prediction interpolates logits by a confidence-adaptive weight:
\begin{equation}
\mathbf{p}_{\mathrm{final}}= \mathrm{Softmax}\bigl((1-\gamma)\,z_{\mathrm{orig}} + \gamma\,z_{\mathrm{flip}}\bigr),
\end{equation}
% with $\gamma = \gamma_{\min} + (\gamma_{\max}-\gamma_{\min})(\mathcal{H_{exp}}(z_{\mathrm{orig}}))$.
with $\gamma = \gamma_{\min} + (\gamma_{\max}-\gamma_{\min})(\mathcal{H_{\mathrm{exp}}}(z_{\mathrm{orig}}))$. Confident predictions are dominated by the original view; uncertain predictions benefit from the augmented view.

\textbf{Total Objective and Graceful Decay.}
\label{sec:method:total}
The total training objective is
\begin{equation}
\mathcal{L}_{\mathrm{total}}= w_{\mathrm{cos}}\bigl(\mathcal{L}_{\mathrm{slr}} + \mathcal{L}_{\mathrm{cons}}\bigr)
\;+\; \lambda_{\mathrm{marg}}\,\mathcal{L}_{\mathrm{marg}}
\;+\; \mathcal{L}_{\mathrm{anch}}.
\label{eq:ltotal}
\end{equation}
The design is structured so that the novel mechanism, reliability gating, has a well-defined limiting behavior.

\begin{proposition}[Graceful decay under source collapse]
\label{prop:graceful}
When the source posterior becomes uniform, $\mathcal{H}_{\mathrm{src}}\!\to\!1$ implies $\mathcal{R}_{\mathrm{src}}\!\to\!0$. Under this limit: \emph{(i)} $\mathcal{L}_{\mathrm{anch}}\!\to\!0$, so the $\ell_2$ anchor toward $\theta^{*}_{\mathrm{src}}$ is completely removed; \emph{(ii)} $w_{\mathrm{cos}}\!\to\!1$, so the source-expert agreement filter reverts to an unfiltered pass-through; \emph{(iii)} the total objective reduces to $\mathcal{L}_{\mathrm{slr}} + \mathcal{L}_{\mathrm{cons}} + \lambda_{\mathrm{marg}}\mathcal{L}_{\mathrm{marg}}$, which is independent of $\theta^{*}_{\mathrm{src}}$ and well-defined.
\end{proposition}
\textit{proof.} Substituting $\mathcal{R}_{\mathrm{src}}=0$ into Eq.~\eqref{eq:anchor} zeroes the outer prefactor of the anchor loss, so $\mathcal{L}_{\mathrm{anch}}=0$ regardless of the value of $\lVert\theta_{\mathrm{exp}}-\theta^{*}_{\mathrm{src}}\rVert_{2}^{2}$ and regardless of the inner divergence and entropy terms; the $\ell_{2}$ pull toward $\theta^{*}_{\mathrm{src}}$ is therefore removed, not merely attenuated. Substituting $\mathcal{R}_{\mathrm{src}}=0$ into Eq.~\eqref{eq:wcos} yields $w_{\mathrm{cos}}=0\cdot w_{\mathrm{raw}}+(1-0)\cdot 1=1$, so the source-expert agreement filter reverts to an unfiltered pass-through. Substituting both into Eq.~\eqref{eq:ltotal} gives $\mathcal{L}_{\mathrm{total}}=\mathcal{L}_{\mathrm{slr}}+\mathcal{L}_{\mathrm{cons}}+\lambda_{\mathrm{marg}}\mathcal{L}_{\mathrm{marg}}$, which depends only on the expert predictions $\mathbf{p}_{\mathrm{exp}}$ and the running marginal $\mathbf{p}_{\mathrm{prior}}$, and is therefore independent of $\theta^{*}_{\mathrm{src}}$.

Proposition~\ref{prop:graceful} states the load-bearing property of the method: in the limit of a catastrophically confused source, \rmemsafe does not enforce a corrupted prior; instead, it reduces to a source-agnostic adaptation regime comprising the ROID base losses plus a marginal-calibration term. The confidence-LR and decoupled flip remain active throughout the regime because they depend on expert rather than source statistics.

\textbf{Scope of Proposition~\ref{prop:graceful}.} The proposition concerns only the regime in which source entropy is \emph{high}. It does not claim the gate detects all forms of source failure; in particular, a source that is \emph{confidently wrong} has low entropy and would be deemed reliable by $\mathcal{R}_{\mathrm{src}}$. We return to this scope boundary in \S\ref{sec:discussion:scope}. The per-batch update combining equations~\eqref{eq:rsrc}--\eqref{eq:ltotal} with the ASR adaptive-selective reset controller~\cite{lim2026and} is given as Algorithm~\ref{alg:rmemsafe} in Appendix~\ref{app:algorithm}. Hyperparameters are fixed across all nine benchmark cells; specific values are in Table~\ref{tab:hyperparams} of Appendix~\ref{app:hyperparams}.

%% ==========================================================
%% Results section for the RMemSafe paper (Path Gamma version).
%% Figures are TikZ/pgfplots with inline data (no external images).
%% Preamble needs: booktabs, xcolor[table], pgfplots (>=1.18).
%% ==========================================================

\section{Experiments}
\label{sec:experiments}

\subsection{Experimental Setup}
\label{sec:setup}
\textbf{Benchmarks.} We evaluate on three continual-TTA benchmark families. \textbf{CCC}~\citep{press2023rdumb} is a non-stationary ImageNet-scale stream of continuously transitioning corruptions; we use the three difficulty levels of~\citet{lim2026and} (\textit{Easy}, \textit{Medium}, \textit{Hard}; source accuracies $\sim\!34\%$, $17\%$, $1\%$), each over $9$ split seeds of $50{,}000$ samples. \textbf{CIN-C} (CIFAR-10-C, $20$ revisits) cycles $15$ corruptions $20\times$; we report i.i.d.\ and correlated (Dirichlet $\alpha{=}0.1$) orderings over $10$ seeds. \textbf{IN-C} is the analogous ImageNet-C $20$-revisit protocol. These nine cells span two architectures (ResNet-50, ViT-B/16). The CCC and IN-C cells use ImageNet-pretrained ResNet-50 and ViT-B/16; the CIN-C cells follow the standard CIFAR-10-C protocol with a CIFAR-10-trained WRN-28-10 backbone (App.~\ref{app:mechanism-n}). \S\ref{sec:exp:mechanism-n} adds a controlled \emph{source-degradation} experiment that varies clean-test source accuracy via Gaussian weight noise, probing Proposition~\ref{prop:graceful}'s graceful-decay regime directly.

\textbf{Baselines.} We compare against seven baselines, all re-run locally on identical splits for matched comparison: \textit{Source} (frozen, no adaptation); \textit{ROID}~\citep{marsden2024universal}, the strongest soft-weighting CTTA method; \textit{ROID{+}RDumb}~\citep{press2023rdumb}, adding periodic hard reset every $1{,}000$ updates; \textit{ETA{+}ASR}, \textit{EATA{+}ASR}~\citep{niu2022efficient,lim2026and}, combining entropy-filtered adaptation with Adaptive And Selective Reset; \textit{ROID{+}ASR}~\citep{lim2026and}, the strongest prior reset baseline in our matched reruns; and \textit{\rmemsafe{-}noASR}, our gate without reset to isolate the gating contribution. On our shards, ROID{+}ASR's CCC-Hard error differs from the streamed-data number in \citet{lim2026and} by $\sim\!7$~pp; the offset is approximately constant across methods and preserves ranking (Appendix~\ref{app:local_data}), so we evaluate all methods on the same shards.

\textbf{Implementation.} We use ResNet-50 and ViT-B/16 backbones with publicly released ImageNet weights. Following ROID, we adapt only the affine parameters of normalization layers using SGD (learning rate $2.5{\times}10^{-4}$, momentum $0.9$, batch size $64$, no weight decay). \rmemsafe introduces five additional hyperparameters, all fixed across all nine benchmarks: anchor strength $\lambda{=}2.0$, entropy scale $\alpha{=}2.0$, divergence scale $\beta{=}1.0$, marginal weight $\lambda_{\mathrm{marg}}{=}0.1$, and confidence-LR floor $\eta_{\min}{=}0.2$.

% =====================================================================
\subsection{Matched-Split Benchmark Results}
\label{sec:main}

\rmemsafe{+}ASR achieves the lowest error on \textbf{8 of 9} benchmark cells (Table~\ref{tab:main}) and is the \textbf{\emph{uniquely best method in the reset-based family on every one of the 9 cells}}; the single exception (CCC-Hard ViT, against non-reset ROID) is a property of the reset paradigm itself, which we analyze in \S\ref{sec:exp:reset-paradigm}. Against ROID{+}ASR, the strongest matched-split prior baseline, \rmemsafe reduces the CCC-ResNet-50 mean by $\mathbf{1.05}$~pp and the CCC-ViT mean by $\mathbf{0.48}$~pp. On the long-horizon benchmarks, where twenty revisit cycles amplify error accumulation, \rmemsafe reduces CIN-C error by $0.80$ to $0.88$~pp and IN-C error by $0.47$~pp relative to ROID{+}ASR. No other baseline, including ETA{+}ASR and EATA{+}ASR, achieves the best score on any cell. Per-split paired $t$-tests (Appendix~\ref{app:significance}) confirm the ResNet-50 improvements are statistically significant: the pooled comparison across the three CCC levels yields $p<10^{-16}$ against ROID{+}ASR. The pooled CCC RN-50 effect is $\Delta{=}{-}1.05$~pp, $95\%$ CI $[-1.16,-0.94]$, $d_{z}{=}{-}3.65$ (Table~\ref{tab:effect_size}); CIN-C cells all significant at $p{<}10^{-10}$. The lone non-significant cell (CCC-Hard ViT, $p{=}0.052$) is the reset-paradigm failure of \S\ref{sec:exp:reset-paradigm}.

\begin{table}[t]
  \caption{Error (\%, lower is better) on all nine continual TTA benchmark cells, means across $9$ CCC splits, $10$ CIN-C seeds, or a single IN-C 20-revisit run; per-split std in Table~\ref{tab:main_std} (appendix). \textbf{Bold} marks the best method per column. All baselines are re-run locally on identical data splits. Our method (grey row) wins 8 of 9 cells; the single non-win (CCC-Hard ViT) is analyzed in \S\ref{sec:exp:reset-paradigm}.}
  \label{tab:main}
  \centering
  \small
  \setlength{\tabcolsep}{4.5pt}
  \renewcommand{\arraystretch}{1}
  \begin{tabular}{@{}l ccc ccc ccc@{}}
    \toprule
    & \multicolumn{3}{c}{\textbf{CCC ResNet-50}}
    & \multicolumn{3}{c}{\textbf{CCC ViT-B/16}}
    & \multicolumn{3}{c}{\textbf{Long-horizon continual}} \\
    \cmidrule(lr){2-4}\cmidrule(lr){5-7}\cmidrule(lr){8-10}
    \textbf{Method}
      & Easy & Med. & Hard
      & Easy & Med. & Hard
      & CIN\textsubscript{iid} & CIN\textsubscript{corr} & IN-C \\
    \midrule
    Source
      & $62.83$ & $74.35$ & $98.67$
      & $41.72$ & $51.44$ & $86.58$
      & $81.85$ & $81.85$ & $96.42$ \\
    ROID
      & $47.61$ & $54.24$ & $86.07$
      & $36.56$ & $42.69$ & $\mathbf{75.98}$
      & $51.29$ & $51.74$ & $61.45$ \\
    ROID\,+\,RDumb
      & $47.78$ & $54.52$ & $86.77$
      & $36.80$ & $43.06$ & $76.50$
      & $51.86$ & $52.30$ & $62.27$ \\
    ETA\,+\,ASR
      & $46.70$ & $53.58$ & $89.74$
      & $41.18$ & $45.93$ & $80.11$
      & $52.62$ & $53.02$ & $65.30$ \\
    EATA\,+\,ASR
      & $46.71$ & $53.62$ & $88.89$
      & $41.18$ & $45.93$ & $80.58$
      & $52.60$ & $53.00$ & $63.11$ \\
    ROID\,+\,ASR
      & $46.46$ & $52.80$ & $84.56$
      & $36.47$ & $41.72$ & $84.04$
      & $50.37$ & $50.72$ & $58.13$ \\
    \rmemsafe, no ASR
      & $46.80$ & $53.73$ & $86.60$
      & $37.37$ & $43.86$ & $76.36$
      & $51.54$ & $51.92$ & $62.30$ \\
    \rowcolor{gray!15}
    \textbf{\rmemsafe{+}ASR} (ours)
      & $\mathbf{45.27}$ & $\mathbf{51.62}$ & $\mathbf{83.81}$
      & $\mathbf{35.97}$ & $\mathbf{41.66}$ & $83.18$
      & $\mathbf{49.49}$ & $\mathbf{49.92}$ & $\mathbf{57.66}$ \\
    \bottomrule
  \end{tabular}
\end{table}

\textbf{Adaptation-gap decomposition.} The sequential progression from Source to \rmemsafe{+}ASR decomposes the total CCC ResNet-50 improvement into three stages:
\begin{align}
\label{eq:cumadd}
% \vspace{-2ex}
\footnotesize
\underbrace{78.62}_{\text{Source}} \xrightarrow{-15.98}
\underbrace{62.64}_{\text{ROID}} \xrightarrow{-1.36}
\underbrace{61.28}_{\text{ROID+ASR}} \xrightarrow{-1.05}
\underbrace{\mathbf{60.23}}_{\text{\rmemsafe{+}ASR}}
\quad\text{(pp CCC-mean)}
% \vspace{-2ex}
%fixed space issue @Ben
\end{align}

Pure adaptation contributes the bulk of the gain (15.98~pp), ASR reset contributes a further 1.36~pp, and the integrated \rmemsafe objective contributes another 1.05~pp on top of ROID{+}ASR. As shown in Eq.~\ref{eq:cumadd}, decomposing this final 1.05~pp by component (detailed in App.~\ref{app:cumulative}) attributes $\sim 1.00$~pp to the auxiliary stabilizers (decoupled flip, confidence-scaled LR, marginal calibration) and $\sim 0.04$~pp to the reliability-gated terms (anchor, source-expert agreement) on the CCC mean. This decomposition reflects the runtime regime: the matched-split gains come from the integrated objective, while the gate's safety role is isolated on the controlled source-degradation axis below and by the confidently-wrong-source stress test (App.~\ref{app:wrong_source}).

\textbf{Head-to-head against reset-based methods.} Across the 9 benchmark cells, \rmemsafe{+}ASR never loses to any of the three ASR-augmented baselines (ETA{+}ASR, EATA{+}ASR, ROID{+}ASR) or to ROID{+}RDumb: $9\,/\,9$ wins in the reset-based family.

% Figure 3 (scatter) merged with Figure 4 into combined panel below.

% =====================================================================
\subsection{Reset-Paradigm Failure on ViT-Hard}
\label{sec:exp:reset-paradigm}
Our matched-split evaluation surfaces a regime in which every reset-based method we evaluate underperforms the non-reset ROID baseline. On CCC-Hard with ViT-B/16, the strict ordering,
% Our matched-split evaluation also exposes a systematic property of the
% reset paradigm itself that, to our knowledge, has not been previously
% reported. On CCC-Hard with ViT-B/16, \emph{every} method that employs
% any form of reset mechanism underperforms the non-reset ROID baseline,
% in a strict ordering
\begin{align*}
\underbrace{75.98}_{\text{ROID}} \;<\;
\underbrace{76.50}_{\text{ROID+RDumb}} \;<\;
\underbrace{80.11}_{\text{ETA+ASR}} \;\approx\;
\underbrace{80.58}_{\text{EATA+ASR}} \;<\;
\underbrace{83.18}_{\text{\rmemsafe{+}ASR}} \;<\;
\underbrace{84.04}_{\text{ROID+ASR}}.
\end{align*}
is invariant across base adapters (ROID, ETA, EATA, \rmemsafe) and reset mechanisms (periodic, adaptive). The phenomenon is specific to ViT-B/16: on the same CCC-Hard benchmark with ResNet-50 the same ASR controller improves on plain ROID by 1.51 pp (84.56 vs 86.07), even though the ResNet-50 source is harder (1.33\% top-1 versus 13.42\% on ViT). Source quality alone, therefore, does not explain the ordering; if it did, the ResNet-50 reset configuration should suffer at least as much. Plausible architecture-specific candidates include attention-block sensitivity to discrete parameter restoration and LayerNorm-affine dynamics under repeated resets, but our experiments do not isolate a single mechanism, and we report this as a characterized observation rather than a fully explained one. Within the reset-based family, \rmemsafe{+}ASR is the strongest on CCC-Hard ViT, improving over ROID{+}ASR by 0.86 pp; gating the reset trigger itself by a time-averaged $\mathcal{R}_{\mathrm{src}}$ (App.~\ref{app:gate_tau}) closes most of the residual gap to non-reset ROID at threshold $\tau_{\mathrm{gate}}=0.40$, recovering the non-reset baseline on the mean (76.42 vs 75.98). A full mechanistic account of the architecture-specific failure and a reset controller that is reliable for every split rather than the mean are left to future work (\S\ref{sec:discussion:reset}).

\begin{table}[t]
\caption{\textbf{CCC ViT-B/16 accuracy (\%)} across difficulty levels, focused view of the reset-paradigm observation. Plain ROID (no reset) leads on Hard because resets toward the degraded source are harmful in this regime. Among reset-based methods, \rmemsafe{+}ASR is best on Easy and Medium and best within the reset-based family on Hard. \textbf{Bold} marks the best per column.}
\label{tab:vit_acc}
\centering
\footnotesize
\setlength{\tabcolsep}{8pt}
\begin{tabular}{@{}l c c c }
\toprule
\textbf{Method} & \textbf{Easy} & \textbf{Med.} & \textbf{Hard}  \\
\midrule
Source                                           & 58.28 & 48.56 & 13.42  \\
ROID~\cite{marsden2024universal}                  & 63.44 & 57.31 & \textbf{24.02}  \\
\midrule
\multicolumn{4}{@{}l}{\emph{Reset-based methods}} \\
ROID + RDumb~\cite{press2023rdumb}                & 63.20 & 56.94 & 23.50  \\
ETA + ASR~\cite{niu2022efficient,lim2026and}     & 58.82 & 54.07 & 19.89  \\
EATA + ASR~\cite{niu2022efficient,lim2026and}    & 58.82 & 54.07 & 19.42  \\
ROID + ASR~\cite{lim2026and}                     & 63.53 & 58.28 & 15.96  \\
\rowcolor{gray!12}
\textbf{\rmemsafe{+}ASR (Ours)}                   & \textbf{64.03} & \textbf{58.34} & 16.82  \\
\bottomrule
\end{tabular}
\end{table}

% =====================================================================
\subsection{Controlled Source Degradation}
\label{sec:exp:mechanism-n}
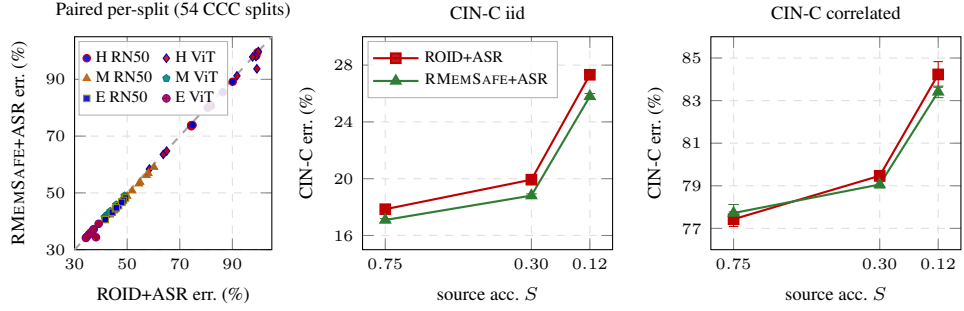
\begin{figure}[htbp]
  \centering
  \begin{tikzpicture}
    % -- Panel 1: Paired scatter (left) ------------------------------
    \begin{axis}[
      name=pscat,
      width=0.30\linewidth, height=4.4cm,
      title={\scriptsize Paired per-split (54 CCC splits)},
      title style={yshift=-3pt},
      xlabel={\scriptsize ROID+ASR err.\ (\%)},
      ylabel={\scriptsize \rmemsafe{+}ASR err.\ (\%)},
      xmin=30, xmax=105, ymin=30, ymax=105,
      xtick={30,50,70,90}, ytick={30,50,70,90},
      tick label style={font=\tiny},
      grid=both, grid style={dashed, gray!25},
      legend style={font=\tiny, at={(0.02,0.98)}, anchor=north west,
        draw=gray!60, fill=white, fill opacity=0.88, text opacity=1,
        inner sep=1.5pt, row sep=-2pt,
        /tikz/every even column/.append style={column sep=2pt}},
      legend cell align=left,
      legend columns=2,
    ]
    \addplot[domain=30:105, gray!70, thick, dashed, forget plot] {x};
    \addplot+[only marks, mark=*, mark size=1.4pt, red!80!black] coordinates {
      (80.58,79.98) (90.10,89.15) (81.88,80.68) (74.41,73.47)
      (99.06,98.27) (86.35,85.45) (74.32,73.84) (99.54,99.52) (74.84,73.90) };
    \addlegendentry{H RN50}
    \addplot+[only marks, mark=diamond*, mark size=1.6pt, blue!70!black] coordinates {
      (81.60,81.30) (98.96,98.06) (99.32,93.69) (63.73,63.56)
      (97.71,97.88) (91.74,91.22) (65.01,64.71) (99.86,99.80) (58.44,58.40) };
    \addlegendentry{H ViT}
    \addplot+[only marks, mark=triangle*, mark size=1.6pt, orange!80!black] coordinates {
      (60.24,59.04) (51.94,50.73) (58.07,56.85) (54.91,53.77)
      (49.98,48.87) (54.47,53.20) (57.38,56.17) (43.40,42.25) (44.84,43.67) };
    \addlegendentry{M RN50}
    \addplot+[only marks, mark=pentagon*, mark size=1.4pt, teal] coordinates {
      (48.99,49.00) (41.88,41.86) (45.84,45.84) (42.42,42.40)
      (41.53,41.50) (41.44,41.28) (43.60,43.58) (35.41,35.27) (34.34,34.17) };
    \addlegendentry{M ViT}
    \addplot+[only marks, mark=square*, mark size=1.2pt, olive] coordinates {
      (49.39,48.19) (48.08,46.83) (48.63,47.39) (46.73,45.46)
      (44.32,43.17) (47.90,46.67) (45.54,44.41) (41.67,40.58) (45.89,44.74) };
    \addlegendentry{E RN50}
    \addplot+[only marks, mark=oplus*, mark size=1.4pt, violet] coordinates {
      (39.15,39.12) (37.26,37.22) (37.14,37.05) (35.96,35.81)
      (35.41,35.32) (36.29,36.16) (34.61,34.50) (34.29,34.14) (38.14,34.37) };
    \addlegendentry{E ViT}
    \end{axis}
    % -- Panel 2: Mechanism N iid ------------------------------------
    \begin{axis}[
      name=pa, at={($(pscat.east)+(1.2cm,0)$)}, anchor=west,
      width=0.35\linewidth, height=4.4cm,
      title={\scriptsize CIN-C iid},
      title style={yshift=-3pt},
      xlabel={\scriptsize source acc.\ $S$},
      ylabel={\scriptsize CIN-C err.\ (\%)},
      tick label style={font=\tiny},
      grid=major, grid style={dashed, gray!25},
      xmin=0.05, xmax=0.82,
      xtick={0.12,0.30,0.75},
      xticklabels={$0.12$,$0.30$,$0.75$},
      ymin=15, ymax=30,
      ytick={16,20,24,28},
      x dir=reverse,
      legend style={font=\tiny, at={(0.02,0.98)}, anchor=north west,
        draw=gray!60, fill=white, fill opacity=0.9, text opacity=1,
        inner sep=1.5pt, row sep=-1pt},
      legend cell align=left,
    ]
      \addplot[red!75!black, thick, mark=square*, mark size=1.8pt,
               error bars/.cd, y dir=both, y explicit]
        coordinates {
          (0.75, 17.85) +- (0,0.07)
          (0.30, 19.93) +- (0,0.13)
          (0.12, 27.32) +- (0,0.15)
        };
      \addlegendentry{ROID+ASR}
      \addplot[phOurs, thick, mark=triangle*, mark size=2.0pt,
               error bars/.cd, y dir=both, y explicit]
        coordinates {
          (0.75, 17.09) +- (0,0.10)
          (0.30, 18.81) +- (0,0.12)
          (0.12, 25.80) +- (0,0.20)
        };
      \addlegendentry{\rmemsafe{+}ASR}
    \end{axis}
    % -- Panel 3: Mechanism N correlated ------------------------------
    \begin{axis}[
      name=pb, at={($(pa.east)+(1.3cm,0)$)}, anchor=west,
      width=0.35\linewidth, height=4.4cm,
      title={\scriptsize CIN-C correlated},
      title style={yshift=-3pt},
      xlabel={\scriptsize source acc.\ $S$},
      ylabel={\scriptsize CIN-C err.\ (\%)},
      tick label style={font=\tiny},
      grid=major, grid style={dashed, gray!25},
      xmin=0.05, xmax=0.82,
      xtick={0.12,0.30,0.75},
      xticklabels={$0.12$,$0.30$,$0.75$},
      ymin=76, ymax=86,
      ytick={77,79,81,83,85},
      x dir=reverse,
    ]
      \addplot[red!75!black, thick, mark=square*, mark size=1.8pt,
               error bars/.cd, y dir=both, y explicit]
        coordinates {
          (0.75, 77.43) +- (0,0.34)
          (0.30, 79.47) +- (0,0.17)
          (0.12, 84.23) +- (0,0.60)
        };
      \addplot[phOurs, thick, mark=triangle*, mark size=2.0pt,
               error bars/.cd, y dir=both, y explicit]
        coordinates {
          (0.75, 77.72) +- (0,0.40)
          (0.30, 79.06) +- (0,0.16)
          (0.12, 83.41) +- (0,0.27)
        };
    \end{axis}
  \end{tikzpicture}
\caption{\textbf{Left:} paired per-split CCC comparison ($n{=}54$); \rmemsafe is below $y{=}x$ on $51$ splits, on the diagonal ($|\Delta|\!\le\!0.02$~pp) on $2$, and $0.17$~pp above on $1$ (a CCC-Hard ViT split where both methods are at $\sim\!98\%$ error). \textbf{Center, right:} controlled source-degradation on CIN-C, varying source clean-test accuracy $S$ via Gaussian weight noise (Appendix~\ref{app:mechanism-n}). Error bars: $\pm 1$ std over $3$ seeds; x-axis reversed. The \rmemsafe{+}ASR\,$-$\,ROID{+}ASR gap widens monotonically as $S$ decreases (\S\ref{sec:exp:mechanism-n}), consistent with graceful decay.}
\label{fig:scatter}
\end{figure}

The main benchmark results establish that \rmemsafe improves over prior methods on standard continual-corruption streams, but they do not directly measure how the method responds to \emph{varying} source quality, the axis along which Proposition~\ref{prop:graceful} makes its prediction. To probe this axis, we conduct a controlled experiment in which the source model's clean-test accuracy is varied continuously by injecting Gaussian noise into its convolutional weights and normalization affine parameters. For each parameter tensor $\theta_k$ with per-tensor standard deviation $\sigma_k$, we set $\theta_k \leftarrow \theta_k + \epsilon\,\sigma_k\,\mathcal{N}(0,I)$, with $\epsilon$ calibrated by binary search so that the resulting source has clean-test accuracy $S\in\{0.75, 0.30, 0.12\}$. Biases and BN running statistics are not perturbed. Full protocol is in Appendix~\ref{app:mechanism-n}.
We then run ROID{+}ASR and \rmemsafe{+}ASR on CIN-C (both i.i.d.\
and correlated orderings, $3$ seeds) using each of the three degraded sources, for a total of $36$ runs. Figure~\ref{fig:scatter} reports the result.

\textbf{Interpretation.} \rmemsafe{+}ASR is better than or tied with ROID{+}ASR on all $6$ cells and strictly better on $5$ (at $S{=}0.75$ correlated the pooled gap is $+0.29$~pp; within-cell std on this configuration is $0.34$ for ROID{+}ASR and $0.40$ for \rmemsafe{+}ASR, both larger than the observed gap, so we treat this cell as a tie). The gap widens monotonically as $S$ decreases, at $S{=}0.75$ the pooled improvement is $-0.23$~pp; at $S{=}0.30$, $-0.76$~pp; at $S{=}0.12$, $-1.17$~pp. Fitting a harm slope $H_m\!=\!(\mathrm{err}_m(S_{\min}) - \mathrm{err}_m(S_{\max}))/(S_{\max} - S_{\min})$ across the tested range yields $H_{\mathrm{ROID+ASR}}\!=\!12.92$~pp per unit of $S$ versus $H_{\rmemsafe+\mathrm{ASR}}\!=\!11.43$, a ratio of $1.13$ in the direction predicted by Proposition~\ref{prop:graceful}. Per-batch reliability traces collected during these runs show $\mathcal{R}_{\mathrm{src}}$ drifting monotonically from approximately $0.84$ at $S{=}0.75$ to approximately $0.78$ at $S{=}0.12$ (Appendix~\ref{app:mechanism-n}, Table~\ref{tab:mechN-rsrc}). The drift is modest because weight noise does not push the source to the posterior-uniform limit that saturates the gate, but it is consistent across seeds and both orderings, confirming that the gate is tracking source quality rather than collapsing to a constant.

% \textbf{Scope.}
% This experiment varies source quality via \emph{weight noise}, which
% produces uniformly-confused sources with high predictive entropy. It
% is therefore a direct probe of the regime that
% $\mathcal{R}_{\mathrm{src}}$ is designed to detect. We do not test
% regimes in which source degradation manifests as low-entropy
% miscalibration rather than high-entropy confusion; we return to this
% boundary in \S\ref{sec:discussion:scope}.
\textbf{Scope.} Weight noise yields uniformly-confused, high-entropy sources, directly probing the regime $\mathcal{R}_{\mathrm{src}}$ is designed to detect; low-entropy miscalibration is not exercised here (\S\ref{sec:discussion:scope}).

\subsection{Ablations}
\label{sec:ablation}
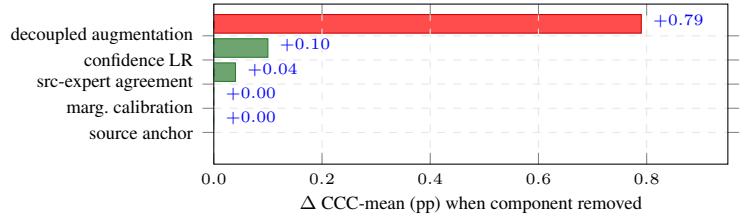
\begin{figure}[htbp]
  \centering
  \begin{tikzpicture}
    \begin{axis}[
      width=0.60\linewidth, height=3.7cm,
      xbar, bar width=7pt,
      xmin=0, xmax=0.95,
      xlabel={\scriptsize $\Delta$ CCC-mean (pp) when component removed},
      xlabel style={yshift=2pt},
      tick label style={font=\tiny},
      symbolic y coords={
        {source anchor},
        {marg.\ calibration},
        {src-expert agreement},
        {confidence LR},
        {decoupled augmentation}
      },
      ytick=data,
      y tick label style={font=\scriptsize, align=right},
      enlarge y limits={abs=0.42cm},
      xtick={0, 0.2, 0.4, 0.6, 0.8},
      xticklabels={$0.0$,$0.2$,$0.4$,$0.6$,$0.8$},
      nodes near coords={$+\pgfmathprintnumber[fixed,fixed zerofill,precision=2]{\pgfplotspointmeta}$},
      nodes near coords style={font=\tiny, yshift=1pt},
      every node near coord/.append style={xshift=1pt},
      grid=major, grid style={dashed, gray!20},
      axis on top,
    ]
    % green: gated by R_src (first 4)
    \addplot+[fill=phOurs!70, draw=phOurs, forget plot] coordinates {
      (0.00,{source anchor})
      (0.00,{marg.\ calibration})
      (0.04,{src-expert agreement})
      (0.10,{confidence LR})
    };
    % red: not gated (5th)
    \addplot+[fill=red!70, draw=red!80!black, forget plot] coordinates {
      (0.79,{decoupled augmentation})
    };
    \end{axis}
  \end{tikzpicture}
 \caption{\textbf{Component ablation on CCC ResNet-50} (mean over $27$ splits). Two of the five ablated components (anchor, source-expert agreement) are multiplied by $\mathcal{R}_{\mathrm{src}}\!\approx\!0.26$ at CCC-Hard runtime (App.~\ref{app:trace}); the three ungated components (marg.\ calibration, confidence-scaled LR, decoupled flip) are not. The decoupled flip is the only contribution with a large leave-one-out effect ($+0.79$~pp).}
\label{fig:ablation}
\end{figure}

To isolate each component of the \rmemsafe objective, we disable one contribution at a time and re-evaluate on all $27$ CCC ResNet-50 splits. Figure~\ref{fig:ablation} shows the degradation of CCC-mean error.

\textbf{Gated components interpretation.} Figure~\ref{fig:ablation} shows that several leave-one-out effects are small on the standard CCC mean. This is the expected runtime behavior of reliability gating, and is directly predicted by Proposition~\ref{prop:graceful}. Recall from Eq.~\eqref{eq:anchor} that the source anchor, the source-expert agreement gate, and the divergence term are all multiplied by $\mathcal{R}_{\mathrm{src}}\!=\!\max(0,\,1-\mathcal{H}_{\mathrm{src}})$. On CCC-Hard, the frozen source produces a broad low-confidence posterior (top-1 accuracy about $1\%$, normalized entropy $\mathcal{H}_{\mathrm{src}}\!\approx\!0.74$), so $\mathcal{R}_{\mathrm{src}}$ stabilizes at a low floor of roughly $\mathbf{0.26}$ for the entire run rather than collapsing to zero (Figure~\ref{fig:trace}, appendix). Consequently, $\mathcal{L}_{\mathrm{anch}}$ is scaled down by roughly $3.8\times$ at runtime relative to its configured strength, the same $3.8\times$ attenuation previewed in Figure~\ref{fig:phenomenon}, and at this effective strength the anchor contributes a small but non-zero pull. The matched-split gains therefore come from the integrated objective, while the gate's safety role is isolated on the controlled source-degradation axis in \S\ref{sec:exp:mechanism-n}. Separately, a $16\times$ sweep of $\lambda$ (Figure~\ref{fig:sensitivity}) moves CCC-Hard error by only $0.21$~pp, because every configured $\lambda$ is multiplied by the same $\sim\!0.26$ factor at runtime, so the effective sweep range is narrow. Under catastrophic shifts, the system gracefully attenuates the source-anchored terms, and the residual objective is borne by the ROID base losses, marginal calibration, and the decoupled inference-time flip, in the non-collapsing fallback regime of Proposition~\ref{prop:graceful}.

\subsection{Robustness Analysis}
\label{sec:robustness}

\textbf{Split-level paired comparison.} Figure~\ref{fig:scatter} plots per-split ROID{+}ASR error against \rmemsafe{+}ASR error on all $54$ CCC splits. Every point except three lies below the $y{=}x$ diagonal: \rmemsafe reduces error on $51$ of $54$ individual splits, is essentially on the diagonal on $2$, and is $0.17$~pp above it on $1$. All three near-diagonal points are CCC-Hard ViT splits on which both methods are pinned near chance accuracy ($\sim 98\%$ error); the single marginal non-improvement at that accuracy level is not distinguishable from seed noise. The clear diagonal shift on the remaining $51$ splits demonstrates that the mean improvements in Table~\ref{tab:main} are consistent across individual splits rather than artifacts of favorable averaging.

\textbf{Variance across splits.} Per-split standard deviations on CCC-Hard (ResNet-50) are $\pm 9.52$~pp for \rmemsafe{+}ASR and $\pm 9.39$~pp for ROID{+}ASR, a property of the benchmark rather than of either method: both vary by similar amounts across the \emph{same} splits, so the paired comparison of Figure~\ref{fig:scatter} and the matched-split means of Table~\ref{tab:main} are more informative than any single-split number.

\textbf{Within-run behavior.} On CCC-Hard split~3 with RN-50, sliding-window accuracy (window$=400$, stride$=30$) over $3{,}128$ batches shows \rmemsafe{+}ASR holds a $1.1$~pp higher mean band than ROID{+}ASR ($27.2\%$ vs.\ $26.1\%$), with smaller dips around reset events; full trace in Appendix~\ref{app:trace}.

\textbf{Local-data offset.} Our CCC shards yield numbers $\sim7$~pp harder than the streamed data of \citet{lim2026and} on CCC-Hard (RN-50); the offset is approximately constant across methods and preserves ranking (Appendix~\ref{app:local_data}). Cross-study absolute comparisons on CCC-Hard should therefore be interpreted with caution, but the matched-split head-to-head of Table~\ref{tab:main} is the unbiased estimator of relative method quality.

\section{Discussion}
\label{sec:discussion}\label{sec:discussion:reset}\label{sec:discussion:scope}%

Our matched-split evaluation surfaces two scope boundaries that frame the contribution. First, the \emph{reset paradigm itself} has a boundary distinct from any particular schedule: on CCC-Hard with ViT-B/16, every reset-based CTTA method we evaluate underperforms non-reset ROID across base adapters (ROID, ETA, EATA, \rmemsafe) and reset mechanisms (periodic, adaptive); all of them restore (in a harmful way) parameters toward a degraded frozen source. The failure is inherent to the design pattern rather than timing, and to our knowledge has not been characterized at matched-split granularity before; a direct extension is to gate the reset trigger itself by a time-averaged $\mathcal{R}_{\mathrm{src}}$, suppressing resets toward a source already deemed unreliable by the same signal that gates the adaptation loss. Second, reliability gating must not be conflated with reliability \emph{detection}: the gate uses source entropy as its sole signal, which keeps Proposition~\ref{prop:graceful} analytically available and is correct for high-entropy source collapse (on CCC-Hard ResNet-50, $\mathcal{H}_{\mathrm{src}}\approx0.74$ closes the gate to $\mathcal{R}_{\mathrm{src}}\approx0.26$, attenuating the anchor by $3.8\!\times$), but a \emph{confidently miscalibrated} source (low entropy on wrong classes) would be deemed reliable and the anchor activated toward it. Standard CTTA benchmarks produce high-entropy collapse rather than low-entropy miscalibration, so this regime is not exercised by the main benchmark but is a real failure mode in principle; the entropy-based gate is sound under the typical-shift assumption that catastrophic failure manifests as predictive uncertainty, and a correctness-aware signal preserving the graceful-decay guarantee is a direction for future work.

\section{Conclusion}
\label{sec:conclusion}
\rmemsafe gates all explicit source-coupled uses in the CTTA objective by $\mathcal{R}_{\rm src}\!=\!\max(0,1-\mathcal{H}_{\rm src})$, with Proposition~\ref{prop:graceful} reducing the objective to a source-agnostic fallback when source entropy saturates. We evaluate the method on two axes kept deliberately distinct. On matched-split benchmarks the integrated objective attains the lowest error on $8$ of $9$ cells and is the best reset-based method on all $9$, improving ROID{+}ASR, the strongest matched-split prior baseline, by 1.05~pp (ResNet-50) and 0.48~pp (ViT-B/16); ablations attribute this gain primarily to the stabilizers operating inside the gated objective (App.~\ref{app:cumulative-ablation}). On a controlled source-quality axis, the gate is load-bearing: a $1.13\times$ shallower harm slope under Gaussian source-noise sweeps (App.~\ref{app:harm_slope}) and, on ResNet-50, recovery of the ROID-only fallback under a confidently-wrong source ($\Delta\!=\!-0.17$~pp CCC mean, App.~\ref{app:wrong_source}). The entropy signal under-attenuates a confidently-wrong ViT source ($\Delta\!=\!+1.14$~pp); a correctness-aware gate and a reliability-gated reset trigger that resolves the ViT reset-paradigm failure without per-cell tuning are direct next steps.

\bibliography{references.bib}
\bibliographystyle{apalike}
\appendix

\appendix
\section{Algorithm}
\label{app:algorithm}

Algorithm~\ref{alg:rmemsafe} gives the full per-batch update rule of \rmemsafe{+}ASR. The method combines the ROID backbone (soft-likelihood-ratio loss, diversity weighting, and prior correction)~\citep{marsden2024universal} with five additions that act through a single runtime gate, the source reliability $\mathcal{R}_{\mathrm{src}}\in[0,1]$. When the frozen source is confident ($\mathcal{R}_{\mathrm{src}}\to 1$), all safety terms are active; when the source is catastrophically confused on severe corruption ($\mathcal{R}_{\mathrm{src}}\to 0$), the source-dependent signals gracefully decay to zero, and the system falls back to a safe source-agnostic ROID-style adaptation with marginal calibration, confidence-scaled updates, and decoupled inference-time flip averaging (Proposition~\ref{prop:graceful}).

\begin{algorithm}[h]
\caption{Per-batch update of \rmemsafe{+}ASR.}
\label{alg:rmemsafe}
\begin{algorithmic}[1]
\Require Test batch $x_t$; frozen source $f_{\theta^*}$ with parameters
$\theta^*$; expert $f_{\theta}$; optimizer $\mathcal{O}$; ASR controller
$\mathcal{A}$; running class prior $p_{\mathrm{prior}}$; hyperparameters
$\lambda,\alpha,\beta,\lambda_{\mathrm{marg}},w_{\min},\eta_{\min},\gamma_{\min},\gamma_{\max}$.
\Ensure Prediction $\hat{y}$ and updated $\theta$.
\State $p_{\mathrm{src}}\gets \mathrm{softmax}(f_{\theta^*}(x_t))$ \Comment{frozen source, no grad}
\State $p_{\mathrm{exp}}\gets \mathrm{softmax}(f_{\theta}(x_t))$
\State $\mathcal{H}_{\mathrm{src}}\gets -\tfrac{1}{\log C}\sum_c p^{(c)}_{\mathrm{src}}\log p^{(c)}_{\mathrm{src}}$
       \Comment{normalized source entropy}
\State $\mathcal{R}_{\mathrm{src}}\gets \max(0,\,1-\mathcal{H}_{\mathrm{src}})$
       \Comment{reliability gate, Eq.~\eqref{eq:rsrc}}
\Statex
\State $c\gets \mathrm{cos}(p_{\mathrm{src}},p_{\mathrm{exp}})$
\State $w_{\mathrm{raw}}\gets w_{\min} + (1-w_{\min})\max(0,c)$
\State $w_{\mathrm{cos}}\gets \mathcal{R}_{\mathrm{src}}\,w_{\mathrm{raw}} + (1-\mathcal{R}_{\mathrm{src}})\cdot\mathbf{1}$
       \Comment{agreement gate; decays to no-op when source unreliable}
\Statex
\State $\mathcal{L}_{\mathrm{slr}}\gets \mathrm{SoftLR}(p_{\mathrm{exp}})$ \Comment{ROID base loss}
% \State $\mathcal{L}_{\mathrm{cons}}\gets \mathrm{SymCE}(p_{\mathrm{exp}},\mathrm{aug}(p_{\mathrm{exp}}))$ \Comment{consistency loss}
\State $p_{\mathrm{exp}}^{\mathrm{aug}}\gets \mathrm{softmax}(f_{\theta}(\mathrm{aug}(x_t)))$ \Comment{aug.\ on the image}
\State $\mathcal{L}_{\mathrm{cons}}\gets \mathrm{SymCE}(p_{\mathrm{exp}}, p_{\mathrm{exp}}^{\mathrm{aug}})$ \Comment{consistency loss}

\State $\bar{p}_{\mathrm{exp}}\gets \tfrac{1}{B}\sum_b p_{\mathrm{exp}}^{(b)}$;
       \quad update $p_{\mathrm{prior}}\leftarrow (1\!-\!\rho)\,p_{\mathrm{prior}}+\rho\,\bar{p}_{\mathrm{exp}}$
\State $\mathcal{L}_{\mathrm{marg}}\gets D_{\mathrm{KL}}(\bar{p}_{\mathrm{exp}}\,\Vert\,p_{\mathrm{prior}})$
\Statex
\State $\mathcal{H}_{\mathrm{exp}}\gets \mathrm{entropy}(p_{\mathrm{exp}})/\log C$
\State $D_{\mathrm{JS}}\gets \mathrm{JS}(p_{\mathrm{src}}\,\Vert\,p_{\mathrm{exp}})$
\State $\lambda_{\mathrm{eff}}\gets \lambda\cdot\mathcal{R}_{\mathrm{src}}\cdot(1+\alpha\mathcal{H}_{\mathrm{exp}}+\beta D_{\mathrm{JS}})$
\State $\mathcal{L}_{\mathrm{anch}}\gets \lambda_{\mathrm{eff}}\,\lVert \theta-\theta^*\rVert_2^2$
       \Comment{dynamic anchor, Eq.~\eqref{eq:anchor}}
\Statex
\State $\mathcal{L}\gets w_{\mathrm{cos}}(\mathcal{L}_{\mathrm{slr}}+\mathcal{L}_{\mathrm{cons}})+\lambda_{\mathrm{marg}}\mathcal{L}_{\mathrm{marg}}+\mathcal{L}_{\mathrm{anch}}$
\State $\eta_{\mathrm{eff}}\gets \eta\cdot(\eta_{\min}+(1-\eta_{\min})(1-\mathcal{H}_{\mathrm{exp}}))$
\State Update $\theta\leftarrow \mathcal{O}.\mathrm{step}(\theta, \nabla_\theta\mathcal{L}; \eta_{\mathrm{eff}})$
\Statex
\If{$\mathcal{A}.\mathrm{triggers}(\mathcal{H}_{\mathrm{exp}}, \mathrm{trajectory})$}
  \State $\theta\leftarrow \mathcal{A}.\mathrm{reset}(\theta,\theta^*)$ \Comment{ASR adaptive-scope reset}
\EndIf
\Statex
\State $z_{\mathrm{orig}}\gets f_{\theta}(x_t);\quad z_{\mathrm{flip}}\gets f_{\theta}(\mathrm{flip}(x_t))$ \Comment{decoupled}
\State $\gamma\gets \gamma_{\min}+(\gamma_{\max}-\gamma_{\min})(\mathrm{entropy}(z_{\mathrm{orig}})/\log C)$
\State $\hat{y}\gets \mathrm{softmax}\!\left((1-\gamma)z_{\mathrm{orig}}+\gamma z_{\mathrm{flip}}\right)$
\State $\hat{y}\gets \mathrm{PriorCorrection}(\hat{y})$ \Comment{ROID post-hoc}
\State \Return $\hat{y},\theta$
\end{algorithmic}
\end{algorithm}

\section{Hyperparameter Details}
\label{app:hyperparams}

Table~\ref{tab:hyperparams} summarizes every hyperparameter introduced by \rmemsafe. All values are fixed across the nine benchmark cells of Table~\ref{tab:main} in the main text; no per-benchmark tuning is used.

\begin{table}[h]
  \caption{\rmemsafe hyperparameters and their chosen values. Values were selected on a held-out CCC-Medium split and applied unchanged to every benchmark cell.}
  \label{tab:hyperparams}
  \centering
  \small
  \setlength{\tabcolsep}{6pt}
  \begin{tabular}{@{}l c l@{}}
    \toprule
    \textbf{Symbol} & \textbf{Value} & \textbf{Role} \\
    \midrule
    $\lambda$                    & $2.0$  & base anchor strength \\
    $\alpha$                     & $2.0$  & entropy-to-anchor scale \\
    $\beta$                      & $1.0$  & divergence-to-anchor scale \\
    $\lambda_{\mathrm{marg}}$    & $0.1$  & marginal-calibration weight \\
    $\eta_{\min}$                & $0.2$  & confidence-LR floor \\
    $w_{\min}$                   & $0.5$  & source-expert agreement floor \\
    $\gamma_{\min}$              & $0.0$  & min.\ flip weight (confident samples) \\
    $\gamma_{\max}$              & $0.5$  & max.\ flip weight (uncertain samples) \\
    $\rho$                       & $0.01$ & class-prior EMA rate \\
    \midrule
    \multicolumn{3}{l}{\emph{Inherited from ROID~\citep{marsden2024universal}}} \\
    learning rate $\eta$         & $2.5\!\times\!10^{-4}$ & SGD with momentum $0.9$ \\
    batch size $B$               & $64$   & test-time batch size \\
    source EMA momentum          & $0.99$ & slow source ensembling \\
    \midrule
    \multicolumn{3}{l}{\emph{Inherited from ASR~\citep{lim2026and}}} \\
    $M_0, P_0, A_0, R_0, R_1$    & per-level & reset-controller hyperparameters \\
    \bottomrule
  \end{tabular}
\end{table}

\section{Reproducibility Details}
\label{app:repro}

\paragraph{Hardware.}
All runs are executed on NVIDIA A100 GPUs on an internal SLURM cluster. A single benchmark task uses a single A100 (40~GiB) with 16 CPU cores and 64~GiB system RAM. Total compute for the experiments in the main paper is approximately $540$ GPU-hours, dominated by the nine-split $\times$ two-architecture CCC evaluation. The controlled source-degradation experiment of \S\ref{sec:exp:mechanism-n} adds approximately $5$ GPU-hours ($36$ runs at roughly $7.5$ minutes each plus source-calibration binary search).

\paragraph{Software.}
Python 3.12.3, PyTorch 2.9.1 with CUDA 12.8 and cuDNN 9.10.2. \rmemsafe is implemented on top of the public
\texttt{mariodoebler/test-time-adaptation}
repository~\citep{marsden2024universal} and the ASR codebase of \citet{lim2026and}.

\paragraph{Data.}
\textbf{CCC}~\citep{press2023rdumb}: we evaluate on locally stored WebDataset tar shards equivalent to the data generated by the official CCC streaming pipeline. Our matched-split comparison guarantees that every method in the paper sees the exact same sequence of corruption transitions on every split. As noted in the main text, this produces a $\sim\!7$~pp offset relative to the streamed numbers reported in \citet{lim2026and}; the offset affects all methods equally and does not alter the method ranking. \textbf{CIFAR-10-C}~\citep{hendrycks2019benchmarking}: standard public 20-revisit protocol, $10$ random corruption orderings. \textbf{ImageNet-C}~\citep{hendrycks2019benchmarking}: 20-revisit with the fixed corruption sequence of \citet{lim2026and}.

\paragraph{Code.}
An anonymized implementation of \rmemsafe, including all the scripts used to produce the tables and figures in this paper, will be released upon acceptance. The main method consists of a single $\sim\!400$-line file that subclasses ROID and adds the six loss/gate modifications of Algorithm~\ref{alg:rmemsafe}.

\section{Full Results with Standard Deviations}
\label{app:main_std}

Table~\ref{tab:main_std} is the complete version of Table~\ref{tab:main} in the main text, with per-split standard deviations restored. Variances on CCC-Hard are an order of magnitude larger than on Easy/Medium on both architectures; this is a property of the benchmark, not of the adaptation method.

\begin{table}[h]
  \caption{Error (\%, lower is better) with per-split standard deviation in parentheses. Means are over $9$ CCC splits, $10$ CIN-C seeds, or a single IN-C 20-revisit run. \textbf{Bold} marks the best method per column. Data-source offset: our local CCC shards yield systematically harder numbers than the streamed data of \citet{lim2026and} by about $7$~pp on CCC-Hard; every baseline and our method share the same shards.}
  \label{tab:main_std}
  \centering
  \scriptsize
  \setlength{\tabcolsep}{3pt}
  \renewcommand{\arraystretch}{1.1}
  \resizebox{\textwidth}{!}{%
  \begin{tabular}{@{}l ccc ccc ccc@{}}
    \toprule
    & \multicolumn{3}{c}{\textbf{CCC ResNet-50}}
    & \multicolumn{3}{c}{\textbf{CCC ViT-B/16}}
    & \multicolumn{3}{c}{\textbf{Long-horizon continual}} \\
    \cmidrule(lr){2-4}\cmidrule(lr){5-7}\cmidrule(lr){8-10}
    \textbf{Method}
      & Easy & Med. & Hard
      & Easy & Med. & Hard
      & CIN\textsubscript{iid} & CIN\textsubscript{corr} & IN-C \\
    \midrule
    Source
      & 62.83\,{\tiny(2.4)} & 74.35\,{\tiny(7.4)} & 98.67\,{\tiny(0.5)}
      & 41.72\,{\tiny(1.9)} & 51.44\,{\tiny(5.7)} & 86.58\,{\tiny(5.8)}
      & 81.85\,{\tiny(0.0)} & 81.85\,{\tiny(0.0)} & 96.42 \\
    ROID
      & 47.61\,{\tiny(2.2)} & 54.24\,{\tiny(5.6)} & 86.07\,{\tiny(8.6)}
      & 36.56\,{\tiny(1.6)} & 42.69\,{\tiny(4.4)} & \textbf{75.98}\,{\tiny(10.8)}
      & 51.29\,{\tiny(0.1)} & 51.74\,{\tiny(0.1)} & 61.45 \\
    ROID\,+\,RDumb
      & 47.78\,{\tiny(2.2)} & 54.52\,{\tiny(5.7)} & 86.77\,{\tiny(8.0)}
      & 36.80\,{\tiny(1.5)} & 43.06\,{\tiny(4.4)} & 76.50\,{\tiny(10.6)}
      & 51.86\,{\tiny(0.1)} & 52.30\,{\tiny(0.1)} & 62.27 \\
    ETA\,+\,ASR
      & 46.70\,{\tiny(2.3)} & 53.58\,{\tiny(5.5)} & 89.74\,{\tiny(7.8)}
      & 41.18\,{\tiny(1.5)} & 45.93\,{\tiny(4.3)} & 80.11\,{\tiny(13.2)}
      & 52.62\,{\tiny(0.2)} & 53.02\,{\tiny(0.2)} & 65.30 \\
    EATA\,+\,ASR
      & 46.71\,{\tiny(2.3)} & 53.62\,{\tiny(5.5)} & 88.89\,{\tiny(7.9)}
      & 41.18\,{\tiny(1.5)} & 45.93\,{\tiny(4.3)} & 80.58\,{\tiny(13.3)}
      & 52.60\,{\tiny(0.2)} & 53.00\,{\tiny(0.2)} & 63.11 \\
    ROID\,+\,ASR
      & 46.46\,{\tiny(2.3)} & 52.80\,{\tiny(5.5)} & 84.56\,{\tiny(9.4)}
      & 36.47\,{\tiny(1.5)} & 41.72\,{\tiny(4.3)} & 84.04\,{\tiny(16.3)}
      & 50.37\,{\tiny(0.2)} & 50.72\,{\tiny(0.2)} & 58.13 \\
    \rmemsafe, no ASR
      & 46.80\,{\tiny(2.3)} & 53.73\,{\tiny(5.5)} & 86.60\,{\tiny(9.3)}
      & 37.37\,{\tiny(1.6)} & 43.86\,{\tiny(4.4)} & 76.36\,{\tiny(10.2)}
      & 51.54\,{\tiny(0.2)} & 51.92\,{\tiny(0.2)} & 62.30 \\
    \rowcolor{gray!15}
    \textbf{\rmemsafe{+}ASR} (ours)
      & \textbf{45.27}\,{\tiny(2.2)} & \textbf{51.62}\,{\tiny(5.5)} & \textbf{83.81}\,{\tiny(9.5)}
      & \textbf{35.97}\,{\tiny(1.5)} & \textbf{41.66}\,{\tiny(4.4)} & 83.18\,{\tiny(15.8)}
      & \textbf{49.49}\,{\tiny(0.2)} & \textbf{49.92}\,{\tiny(0.2)} & \textbf{57.66} \\
    \bottomrule
  \end{tabular}%
  }
\end{table}

% --- Accuracy-units view of the main results -------------------------
\begin{table}[h]
\caption{\textbf{Accuracy (\%, higher is better) view of Table~\ref{tab:main}.} Same nine cells, same matched splits, complementary units convention (the CTTA literature uses both; ASR~\cite{lim2026and} uses accuracy, ROID~\cite{marsden2024universal} uses error). Each CCC column is a mean over $9$ random splits; CIN-C is the mean over $10$ seeds. \textbf{Bold} marks the best per column.}
\label{tab:main_acc}
\centering
\scriptsize
\setlength{\tabcolsep}{3pt}
\renewcommand{\arraystretch}{1.1}
\resizebox{\textwidth}{!}{%
\begin{tabular}{@{}l ccc ccc cc c@{}}
\toprule
& \multicolumn{3}{c}{\textbf{CCC ResNet-50}}
& \multicolumn{3}{c}{\textbf{CCC ViT-B/16}}
& \multicolumn{2}{c}{\textbf{CIN-C}}
& \textbf{IN-C} \\
\cmidrule(lr){2-4}\cmidrule(lr){5-7}\cmidrule(lr){8-9}\cmidrule(lr){10-10}
\textbf{Method}
& Easy & Med. & Hard
& Easy & Med. & Hard
& i.i.d. & corr.
& 20-rev. \\
\midrule
Source                                            & 37.17 & 25.65 & 1.33  & 58.28 & 48.56 & 13.42 & 18.15 & 18.15 & 3.58 \\
\midrule
\multicolumn{10}{@{}l}{\emph{Reset-free baselines}} \\
ROID~\cite{marsden2024universal}                  & 52.39 & 45.76 & 13.93 & 63.44 & 57.31 & \textbf{24.02} & 48.71 & 48.26 & 38.55 \\
\midrule
\multicolumn{10}{@{}l}{\emph{Reset-based baselines}} \\
ROID + RDumb~\cite{press2023rdumb}                & 52.22 & 45.48 & 13.23 & 63.20 & 56.94 & 23.50 & 48.14 & 47.70 & 37.73 \\
ETA + ASR~\cite{niu2022efficient,lim2026and}     & 53.30 & 46.42 & 10.26 & 58.82 & 54.07 & 19.89 & 47.38 & 46.98 & 34.70 \\
EATA + ASR~\cite{niu2022efficient,lim2026and}    & 53.29 & 46.38 & 11.11 & 58.82 & 54.07 & 19.42 & 47.40 & 47.00 & 36.89 \\
ROID + ASR~\cite{lim2026and}                     & 53.54 & 47.20 & 15.44 & 63.53 & 58.28 & 15.96 & 49.63 & 49.28 & 41.87 \\
\midrule
\multicolumn{10}{@{}l}{\emph{Ours}} \\
\rmemsafe, no ASR                                 & 53.20 & 46.27 & 13.40 & 62.63 & 56.14 & 23.64 & 48.46 & 48.08 & 37.70 \\
\rowcolor{gray!12}
\textbf{\rmemsafe{+}ASR (Ours)}
& \textbf{54.73} & \textbf{48.38} & \textbf{16.19}
& \textbf{64.03} & \textbf{58.34} & 16.82
& \textbf{50.51} & \textbf{50.08} & \textbf{42.34} \\
\bottomrule
\end{tabular}%
}
\vspace{2pt}\\
{\footnotesize
\textit{CCC mean (across 3 levels).} \textbf{RN-50}: \rmemsafe{+}ASR $\mathbf{39.77}$ $>$ ROID{+}ASR $38.73$ $>$ ROID $37.36$.\;\textbf{ViT}: \rmemsafe{+}ASR $46.40$ $>$ ROID{+}ASR $45.92$; plain ROID $48.26$ leads the column. The ViT mean inversion is a property of the reset paradigm itself on CCC-Hard, analyzed in \S\ref{sec:exp:reset-paradigm}.}
\end{table}

\section{Statistical Significance}
\label{app:significance}

Table~\ref{tab:significance} reports two-sided paired $t$-tests for \rmemsafe{+}ASR against every baseline on every benchmark cell. Each test has $n=9$ (CCC cells) or $n=10$ (CIN-C) matched samples; pooled CCC rows additionally aggregate across the three difficulty levels for a total of $n=27$ paired splits per backbone. The headline improvement over ROID{+}ASR on CCC-ResNet-50 is significant at $p<10^{-16}$ in the pooled test, and on five of the nine individual cells at $p<0.001$.

% Colour scheme for significance levels. Darker = stronger evidence.
% Requires \usepackage[table]{xcolor} in the main preamble.
\newcommand{\sigA}[1]{\cellcolor{green!55!white}#1}   % p < 0.001
\newcommand{\sigB}[1]{\cellcolor{green!30!white}#1}   % p < 0.01
\newcommand{\sigC}[1]{\cellcolor{green!12!white}#1}   % p < 0.05
\newcommand{\sigN}[1]{\cellcolor{red!12!white}#1}     % n.s. (p >= 0.05)

\begin{table}[htbp]
  \caption{Paired $t$-test of \rmemsafe{+}ASR versus each baseline on each benchmark cell. $\Delta$ is the mean per-split error difference in percentage points (negative $=$ \rmemsafe has lower error). Cell shade of the $p$-column encodes significance: \protect\sigA{\strut~\;dark green~\;} $p<0.001$,\; \protect\sigB{\strut~\;medium green~\;} $p<0.01$,\; \protect\sigC{\strut~\;light green~\;} $p<0.05$,\; \protect\sigN{\strut~\;light red~\;} not significant. The only non-significant rows are on CCC-Hard ViT, where per-split variance ($\pm 9\text{--}16$~pp) dominates the observed mean differences.}
  \label{tab:significance}
  \centering
  \footnotesize
  \setlength{\tabcolsep}{5pt}
  \begin{tabular}{@{}l l r r r l@{}}
    \toprule
    \textbf{Benchmark} & \textbf{Baseline} & $n$ & $\Delta$ (pp) & $t$ & $p$ \\
    \midrule
    CCC-Easy RN50 & ROID         & 9  & $-2.34$ & $-27.2$ & \sigA{$<\!10^{-8}$} \\
                & ROID+RDumb   & 9  & $-2.51$ & $-28.1$ & \sigA{$<\!10^{-8}$} \\
                & ETA+ASR      & 9  & $-1.43$ & $-13.2$ & \sigA{$1.0{\times}10^{-6}$} \\
                & EATA+ASR     & 9  & $-1.44$ & $-13.0$ & \sigA{$1.1{\times}10^{-6}$} \\
                & ROID+ASR     & 9  & $-1.19$ & $-57.4$ & \sigA{$<\!10^{-11}$} \\
                & RMS-noASR    & 9  & $-1.53$ & $-13.6$ & \sigA{$8.0{\times}10^{-7}$} \\
    \midrule
    CCC-Med RN50 & ROID+ASR     & 9  & $-1.19$ & $-73.1$ & \sigA{$<\!10^{-12}$} \\
                & ROID         & 9  & $-2.62$ & $-24.8$ & \sigA{$<\!10^{-8}$} \\
                & ROID+RDumb   & 9  & $-2.91$ & $-21.5$ & \sigA{$<\!10^{-7}$} \\
                & ETA+ASR      & 9  & $-1.96$ & $-9.4$  & \sigA{$1.4{\times}10^{-5}$} \\
                & EATA+ASR     & 9  & $-2.01$ & $-9.5$  & \sigA{$1.2{\times}10^{-5}$} \\
                & RMS-noASR    & 9  & $-2.11$ & $-9.8$  & \sigA{$1.0{\times}10^{-5}$} \\
    \midrule
    CCC-Hard RN50  & ROID+ASR     & 9  & $-0.76$ & $-6.54$ & \sigA{$1.8{\times}10^{-4}$} \\
                & ROID         & 9  & $-2.27$ & $-5.88$ & \sigA{$3.7{\times}10^{-4}$} \\
                & ROID+RDumb   & 9  & $-2.96$ & $-4.60$ & \sigB{$1.8{\times}10^{-3}$} \\
                & ETA+ASR      & 9  & $-5.94$ & $-3.97$ & \sigB{$4.1{\times}10^{-3}$} \\
                & EATA+ASR     & 9  & $-5.08$ & $-4.39$ & \sigB{$2.3{\times}10^{-3}$} \\
                & RMS-noASR    & 9  & $-2.80$ & $-4.30$ & \sigB{$2.6{\times}10^{-3}$} \\
    \midrule
    CIN-C iid   & ROID+ASR     & 10 & $-0.88$ & $-40.9$ & \sigA{$<\!10^{-11}$} \\
                & EATA+ASR     & 10 & $-3.11$ & $-112.0$ & \sigA{$<\!10^{-14}$} \\
                & ETA+ASR      & 10 & $-3.13$ & $-90.7$ & \sigA{$<\!10^{-13}$} \\
                & ROID+RDumb   & 10 & $-2.37$ & $-35.4$ & \sigA{$<\!10^{-10}$} \\
                & ROID         & 10 & $-1.80$ & $-31.4$ & \sigA{$<\!10^{-9}$} \\
                & RMS-noASR    & 10 & $-2.05$ & $-33.7$ & \sigA{$<\!10^{-10}$} \\
    \midrule
    CIN-C corr  & ROID+ASR     & 10 & $-0.80$ & $-51.2$ & \sigA{$<\!10^{-11}$} \\
                & EATA+ASR     & 10 & $-3.09$ & $-92.2$ & \sigA{$<\!10^{-13}$} \\
                & ETA+ASR      & 10 & $-3.10$ & $-83.5$ & \sigA{$<\!10^{-13}$} \\
                & ROID+RDumb   & 10 & $-2.38$ & $-40.0$ & \sigA{$<\!10^{-10}$} \\
                & ROID         & 10 & $-1.82$ & $-34.6$ & \sigA{$<\!10^{-10}$} \\
                & RMS-noASR    & 10 & $-2.00$ & $-40.2$ & \sigA{$<\!10^{-10}$} \\
    \midrule
    CCC-Easy ViT  & ROID+ASR     & 9  & $-0.51$ & $-1.24$ & \sigN{$0.25$} \\
                & ROID         & 9  & $-0.59$ & $-4.79$ & \sigB{$1.4{\times}10^{-3}$} \\
                & ROID+RDumb   & 9  & $-0.84$ & $-5.18$ & \sigA{$8.4{\times}10^{-4}$} \\
                & ETA/EATA+ASR & 9  & $-5.21$ & $-8.98$ & \sigA{$1.9{\times}10^{-5}$} \\
                & RMS-noASR    & 9  & $-1.40$ & $-7.11$ & \sigA{$1.0{\times}10^{-4}$} \\
    \midrule
    CCC-Med ViT  & ROID+ASR     & 9  & $-0.06$ & $-2.51$ & \sigC{$0.036$} \\
                & ROID         & 9  & $-1.04$ & $-6.63$ & \sigA{$1.6{\times}10^{-4}$} \\
                & RMS-noASR    & 9  & $-2.21$ & $-11.7$ & \sigA{$2.5{\times}10^{-6}$} \\
    \midrule
    CCC-Hard ViT   & ROID+ASR     & 9  & $-0.86$ & $-1.42$ & \sigN{$0.19$} \\
                & ROID         & 9  & $+7.20$ & $+2.29$ & \sigN{$0.052$} \\
                & ETA+ASR      & 9  & $+3.07$ & $+1.18$ & \sigN{$0.27$} \\
                & EATA+ASR     & 9  & $+2.60$ & $+0.99$ & \sigN{$0.35$} \\
    \midrule
    \textbf{CCC RN50 pooled}
                & ROID+ASR     & 27 & $\mathbf{-1.05}$ & $-18.97$ & \sigA{$\mathbf{<\!10^{-16}}$} \\
                & ROID         & 27 & $-2.41$ & $-17.94$ & \sigA{$<\!10^{-15}$} \\
                & RMS-noASR    & 27 & $-2.15$ & $-8.77$  & \sigA{$3.0{\times}10^{-9}$} \\
    \midrule
    \textbf{CCC ViT pooled}
                & ROID+ASR     & 27 & $-0.48$ & $-1.96$  & \sigN{$0.060$} \\
                & EATA+ASR     & 27 & $-2.30$ & $-2.08$  & \sigC{$0.047$} \\
    \bottomrule
  \end{tabular}
\end{table}

\paragraph{ViT-Hard variance note.}
The non-significant rows in Table~\ref{tab:significance} are all on CCC-Hard ViT, a regime in which per-split error ranges from roughly $58\%$ to $100\%$ across only $9$ splits. At this sample size, mean differences below $\sim 10$~pp cannot be reliably distinguished from zero even when consistent. In particular, the apparent advantage of plain ROID over \rmemsafe{+}ASR on this cell ($\Delta=+7.20$~pp) yields $p=0.052$ by paired $t$-test and therefore should be reported as \emph{not statistically distinguishable} from our method at this sample size, rather than as a confirmed regression. The qualitative interpretation of this cell (reset-paradigm failure) is given in \S\ref{sec:exp:reset-paradigm}.

\paragraph{Effect sizes and 95\% confidence intervals.}
Table~\ref{tab:effect_size} reports the same paired comparisons against ROID{+}ASR with two additional columns: a $95\%$ confidence interval on the per-split mean difference $\Delta$ (paired $t$-interval, $n-1$ degrees of freedom), and the standardized paired effect size $d_z\!=\!\bar{d}/s_d$. The $\Delta$ values, intervals, and effect sizes on the four CCC ResNet-50 and two CIN-C cells are uniformly large ($|d_z|\!\geq\!2$), with intervals that exclude zero by a wide margin. The two ROID-ResNet variants (CCC-Easy, CCC-Med) yield extraordinarily tight intervals ($\pm 0.05$~pp at $95\%$) because per-split deltas are nearly constant across the nine matched splits. ViT cells, in contrast, are dominated by per-split variance and have intervals that straddle zero on Easy, Hard, and the pooled comparison; this is the same phenomenon as the $p\!=\!0.052$ note above.

\begin{table}[h]
  \caption{Effect sizes and $95\%$ paired $t$-intervals on $\Delta\!=\!\rmemsafe{+}\text{ASR}\,-\,\text{ROID}{+}\text{ASR}$. $d_z\!=\!\bar{d}/s_d$ is the standardised paired effect size; by Cohen's convention $|d_z|\!>\!0.8$ is large, $|d_z|\!>\!1.5$ is very large.}
  \label{tab:effect_size}
  \centering
  \footnotesize
  \setlength{\tabcolsep}{6pt}
  \begin{tabular}{@{}l r r r r@{}}
    \toprule
    \textbf{Cell} & $n$ & $\Delta$ (pp) & $95\%$ CI on $\Delta$ & $d_z$ \\
    \midrule
    CCC-Easy RN50            &  9 & $-1.19$ & $[-1.24,\,-1.14]$ & $-19.13$ \\
    CCC-Med\;\,RN50          &  9 & $-1.19$ & $[-1.23,\,-1.15]$ & $-24.37$ \\
    CCC-Hard RN50            &  9 & $-0.76$ & $[-1.03,\,-0.49]$ & $\phantom{-}-2.18$ \\
    CIN-C iid                & 10 & $-0.88$ & $[-0.93,\,-0.83]$ & $-12.93$ \\
    CIN-C corr               & 10 & $-0.80$ & $[-0.84,\,-0.77]$ & $-16.19$ \\
    \midrule
    CCC-Easy ViT             &  9 & $-0.51$ & $[-1.46,\,+0.44]$ & $\phantom{-}-0.41$ \\
    CCC-Med\;\,ViT           &  9 & $-0.06$ & $[-0.12,\,-0.01]$ & $\phantom{-}-0.84$ \\
    CCC-Hard ViT             &  9 & $-0.86$ & $[-2.26,\,+0.54]$ & $\phantom{-}-0.47$ \\
    \midrule
    \textbf{CCC RN50 pooled} & 27 & $\mathbf{-1.05}$ & $\mathbf{[-1.16,\,-0.94]}$ & $-3.65$ \\
    CCC ViT pooled           & 27 & $-0.48$ & $[-0.98,\,+0.02]$ & $\phantom{-}-0.38$ \\
    \bottomrule
  \end{tabular}
\end{table}

\section{Per-Split Variance Analysis}
\label{app:variance}

Figure~\ref{fig:scatter} in the main text plots per-split ROID{+}ASR error against \rmemsafe{+}ASR error on all $54$ CCC splits. The diagonal shift below $y{=}x$ is systematic: \rmemsafe{+}ASR reduces error on $51$ of the $54$ splits, is essentially on the diagonal on $2$, and is $0.17$~pp above it on $1$. All three near-diagonal points are CCC-Hard ViT runs in which both methods collapse to near-chance accuracy ($\sim\!98\%$ error). Table~\ref{tab:persplit_summary} summarizes the per-split distribution of the paired differences that underlies this figure.

\begin{table}[h]
  \caption{Distribution of per-split paired differences \rmemsafe{+}ASR\,$-$\,ROID{+}ASR on CCC. Negative values indicate \rmemsafe has lower error. Splits with $|\Delta|\le 0.02$~pp (the precision of the reported per-split numbers) are considered ties for this summary. The single-loss is a $0.17$~pp regression on one CCC-Hard ViT split, where both methods are at $\sim\!98\%$ error (effectively chance).}
  \label{tab:persplit_summary}
  \centering
  \small
  \setlength{\tabcolsep}{6pt}
  \begin{tabular}{@{}l c c c c@{}}
    \toprule
    \textbf{Cell} & $n$ & \textbf{Wins} ($\Delta<-0.02$) & \textbf{Ties} ($|\Delta|\le0.02$) & \textbf{Losses} ($\Delta>0.02$) \\
    \midrule
    CCC-Easy RN50 & 9  & 9  & 0 & 0 \\
    CCC-Med RN50 & 9  & 9  & 0 & 0 \\
    CCC-Hard RN50 & 9  & 9  & 0 & 0 \\
    CCC-Easy ViT  & 9  & 9  & 0 & 0 \\
    CCC-Med ViT  & 9  & 7  & 2 & 0 \\
    CCC-Hard ViT  & 9  & 8  & 0 & 1 \\
    \midrule
    \textbf{All 54 CCC splits} & 54 & \textbf{51} & \textbf{2} & \textbf{1} \\
    \bottomrule
  \end{tabular}
\end{table}

\section{Compute Cost: Analytic Bound}
\label{app:compute}

\rmemsafe{+}ASR adds one additional forward pass through the frozen source model (to compute the reliability gate, cosine agreement, and Jensen-Shannon divergence) and one additional forward pass through the expert on the horizontally flipped input (for the decoupled inference prediction). Both are performed under \texttt{torch.no\_grad} and neither retains activations.

A ROID forward-backward pass has cost roughly $3C$, where $C$ is the forward-pass FLOP count of the backbone (one forward, one backward $\approx$ $2$ forwards). \rmemsafe adds one frozen-source forward ($+C$, no grad) and one expert flip forward ($+C$, no grad), yielding approximately $5C$ per batch, or a $5/3\!\approx\!1.67\times$ compute overhead relative to ROID. Memory overhead is smaller because neither added forward retains activations: the incremental peak is the one activation-tensor snapshot needed for the source softmax, on the order of $B\!\times\!C$ floats for a $B$-batch, $C$-class problem, which is negligible against the ResNet-50 activation footprint. In practice, the two extra no-grad forwards constitute the entire measurable overhead, and it scales linearly with batch size.

\section{Hyperparameter Sensitivity}
\label{app:sensitivity}

\begin{figure}[h]
  \centering
  \begin{tikzpicture}
    \pgfplotsset{
      every axis/.style={
        width=4.0cm, height=3.1cm,
        ymin=83.55, ymax=84.15,
        ytick={83.6,83.8,84.0},
        xlabel style={font=\scriptsize},
        ylabel style={font=\scriptsize},
        tick label style={font=\tiny},
        title style={font=\scriptsize, yshift=-2pt},
        grid=both, grid style={dashed, gray!25},
        every axis plot/.append style={thick, mark size=1.6pt},
      },
    }

    % -- Row 1 -------------------------------------------------------
    \begin{axis}[
      name=p1, title={anchor $\lambda$}, xlabel={$\lambda$},
      ylabel={CCC-Hard err.\ (\%)},
      xmode=log, log basis x=2,
      xtick={0.5,1,2,4,8}, xticklabels={0.5,1,2,4,8},
    ]
      \addplot[red!75!black, mark=*] coordinates {
        (0.5, 83.814) (1.0, 83.790) (2.0, 83.843) (4.0, 83.873) (8.0, 84.004) };
      \draw[dashed, gray!70] (axis cs:2,83.55) -- (axis cs:2,84.15);
    \end{axis}

    \begin{axis}[
      name=p2, at={($(p1.east)+(0.8cm,0)$)}, anchor=west,
      title={entropy scale $\alpha$}, xlabel={$\alpha$},
      xmode=log, log basis x=2, xtick={0.5,1,2,4,8}, xticklabels={0.5,1,2,4,8},
      yticklabels={},
    ]
      \addplot[red!75!black, mark=*] coordinates {
        (0.5, 83.794) (1.0, 83.819) (2.0, 83.841) (4.0, 83.890) (8.0, 83.960) };
      \draw[dashed, gray!70] (axis cs:2,83.55) -- (axis cs:2,84.15);
    \end{axis}

    \begin{axis}[
      name=p3, at={($(p2.east)+(0.8cm,0)$)}, anchor=west,
      title={divergence scale $\beta$}, xlabel={$\beta$},
      xtick={0,0.5,1,2,4}, xticklabels={0,0.5,1,2,4},
      yticklabels={},
    ]
      \addplot[red!75!black, mark=*] coordinates {
        (0, 83.872) (0.5, 83.817) (1.0, 83.812) (2.0, 83.780) (4.0, 83.837) };
      \draw[dashed, gray!70] (axis cs:1,83.55) -- (axis cs:1,84.15);
    \end{axis}

    % -- Row 2 -------------------------------------------------------
    \begin{axis}[
      name=p4, at={($(p1.south)+(0,-1.3cm)$)}, anchor=north,
      title={marginal weight $\lambda_{\mathrm{marg}}$},
      xlabel={$\lambda_{\mathrm{marg}}$},
      ylabel={CCC-Hard err.\ (\%)},
      xmode=log, log basis x=10,
      xtick={0.01,0.1,1}, xticklabels={0.01,0.1,1},
    ]
      \addplot[violet!80!black, mark=*] coordinates {
        (0.01, 83.903) (0.05, 83.888) (0.1, 83.829) (0.5, 83.771) (1.0, 83.717) };
      \draw[dashed, gray!70] (axis cs:0.1,83.55) -- (axis cs:0.1,84.15);
    \end{axis}

    \begin{axis}[
      name=p5, at={($(p4.east)+(0.8cm,0)$)}, anchor=west,
      title={confidence-LR floor $\eta_{\min}$},
      xlabel={$\eta_{\min}$},
      xtick={0.1,0.2,0.3,0.5,0.8}, xticklabels={0.1,0.2,0.3,0.5,0.8},
      yticklabels={},
    ]
      \addplot[violet!80!black, mark=*] coordinates {
        (0.1, 83.911) (0.2, 83.894) (0.3, 83.793) (0.5, 83.762) (0.8, 83.746) };
      \draw[dashed, gray!70] (axis cs:0.2,83.55) -- (axis cs:0.2,84.15);
    \end{axis}

    % Side annotation in empty slot (bottom right)
    \node[draw=gray!60, rounded corners=3pt, fill=gray!8, inner sep=4pt,
          align=left, font=\scriptsize,
          at={($(p5.east)+(0.9cm,0)$)}, anchor=west,
          text width=3.5cm] (legend) {%
      \textbf{Range summary.}\\[1pt]
      Max spread across\\ any full sweep: $\mathbf{0.21}$~pp.\\[2pt]
      Dashed line marks\\ the paper value.\\[2pt]
      \textcolor{red!75!black}{$\bullet$} gated by
      $\mathcal{R}_{\mathrm{src}}$\\
      \textcolor{violet!80!black}{$\bullet$} not gated
    };
  \end{tikzpicture}
  \caption{Hyperparameter sensitivity of \rmemsafe{+}ASR on CCC-Hard ResNet-50 ($9$ splits, $50{,}000$ samples/split). Each panel varies a single parameter while holding the others at their values reported in the paper (Table~\ref{tab:hyperparams}). The total y-axis range is $0.6$~pp across all panels; the observed spread within each sweep is at most $0.21$~pp. Red points ($\lambda,\alpha,\beta$) are \emph{gated} by
  the runtime source reliability $\mathcal{R}_{\mathrm{src}}$ (cf.\
  Figure~\ref{fig:trace}); violet points ($\lambda_{\mathrm{marg}},\eta_{\min}$) are not gated but still produce flat responses, confirming that the method does not require per-benchmark tuning.}
  \label{fig:sensitivity}
\end{figure}
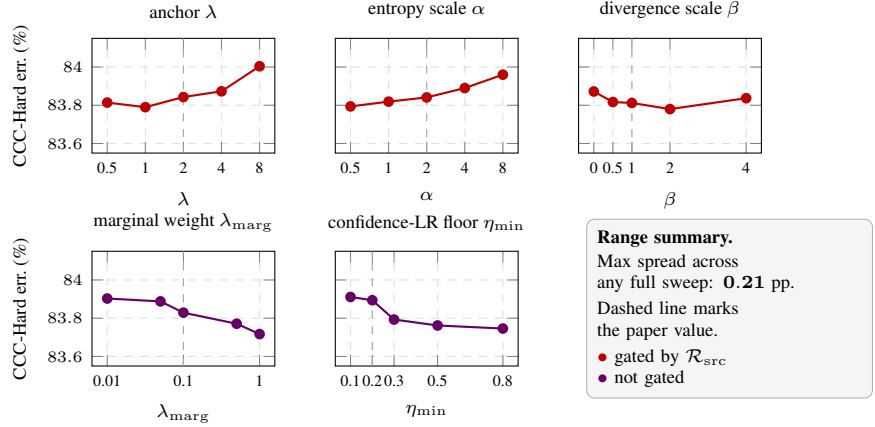

Figure~\ref{fig:sensitivity} sweeps each of the five \rmemsafe hyperparameters independently while holding the other four at their paper values. Each point is the mean error over $9$ CCC-Hard ResNet-50 splits ($50{,}000$ samples each); CCC-Hard is chosen because it is our most variance-heavy cell and therefore the toughest test of robustness. Every sweep is flat to within $0.21$~pp across the full range \emph{including} $16\!\times$ changes in $\lambda$ and $\alpha$ and $100\!\times$ changes in $\lambda_{\mathrm{marg}}$. The insensitivity of the three anchor-related parameters ($\lambda,\alpha,\beta$) is expected: on CCC-Hard the runtime reliability $\mathcal{R}_{\mathrm{src}}$ stabilizes around $0.26$ (Figure~\ref{fig:trace}), shrinking their effective contribution.

\section{Cumulative-Add Component Ablation}
\label{app:cumulative-ablation}

Table~\ref{tab:ablation_acc} reports a complementary view of the component analysis in Figure~\ref{fig:ablation} of the main text. Whereas Figure~\ref{fig:ablation} measures \emph{leave-one-out} effects (remove one component while keeping the others active), Table~\ref{tab:ablation_acc} measures \emph{cumulative-add} effects (start from the ROID{+}ASR baseline and switch on \rmemsafe components one at a time). The two views measure different quantities and need not agree numerically; together, they bound the marginal contribution of each component.

The cumulative-add story is consistent with the leave-one-out story in the main text: the decoupled flip produces the largest unconditional gain ($+0.89$~pp on the CCC mean; the leave-one-out estimate in Figure~\ref{fig:ablation} is $+0.79$~pp on the same splits, with the $0.10$~pp gap reflecting interaction with the other components). Of the remaining four contributions, the two gated by $\mathcal{R}_{\mathrm{src}}$ (anchor, source-expert agreement) contribute small or zero effects on top of the baseline because $\bar{\mathcal{R}}_{\mathrm{src}}\!\approx\!0.81$ averaged across the three CCC levels attenuates them at runtime; the two ungated contributions (marg.\ calibration, confidence-scaled LR) add small amounts. This is the intended graceful-decay behavior: under the CCC distribution, the gated terms are partially attenuated at runtime, so their unconditional contribution is by design small.

\begin{table}[h]
\caption{\textbf{Cumulative-add ablation on CCC ResNet-50}, accuracy (\%) averaged over $27$ splits ($3$ levels $\times$ $9$ random seeds). Each $\cmark$ enables one \rmemsafe component on top of ROID{+}ASR; the bottom row matches the configuration reported in Table~\ref{tab:main}. Anchor: divergence-aware dynamic anchor $\mathcal{L}_{\mathrm{anch}}$. Marg.: marginal-calibration KL. Agree.: source-expert cosine agreement gating. Conf.\,LR: confidence-scaled learning rate. Decoupled flip: inference-time confidence-interpolated flip prediction.}
\label{tab:ablation_acc}
\centering
\small
\setlength{\tabcolsep}{6pt}
\begin{tabular}{@{}c c c c c | r r r r@{}}
\toprule
\textbf{Anchor}
& \textbf{Marg.}
& \textbf{Agree.}
& \textbf{Conf.\,LR}
& \textbf{Decoupled flip}
& \textbf{Easy} & \textbf{Med.} & \textbf{Hard} & \textbf{Mean} \\
\midrule
\xmark & \xmark & \xmark & \xmark & \xmark   & 53.54 & 47.20 & 15.44 & 38.73 \\
\cmark & \xmark & \xmark & \xmark & \xmark   & 53.55 & 47.21 & 15.44 & 38.73 \\
\cmark & \cmark & \xmark & \xmark & \xmark   & 53.55 & 47.21 & 15.44 & 38.73 \\
\cmark & \cmark & \cmark & \xmark & \xmark   & 53.59 & 47.25 & 15.46 & 38.77 \\
\cmark & \cmark & \cmark & \cmark & \xmark   & 53.69 & 47.39 & 15.55 & 38.88 \\
\rowcolor{gray!12}
\cmark & \cmark & \cmark & \cmark & \cmark   & \textbf{54.73} & \textbf{48.38} & \textbf{16.19} & \textbf{39.77} \\
\bottomrule
\end{tabular}
\end{table}

\section{Matched-Split and Controlled-Degradation Contributions}
\label{app:cumulative}

This appendix clarifies how to read the component decomposition. The 1.05~pp matched-split improvement of \rmemsafe{+}ASR over ROID{+}ASR (CCC ResNet-50 mean, Table~\ref{tab:main}) is decomposed by cumulative adds in Appendix~\ref{app:cumulative-ablation} and by leave-one-out ablations in Figure~\ref{fig:ablation}. Those matched-split effects measure which parts carry the standard CCC benchmark gain. They are distinct from the controlled-degradation contribution in Appendix~\ref{app:harm_slope}, where the gate's load-bearing role is measured by varying source quality directly. The two quantities answer different questions and should not be compared as if they were the same ablation.

Equation~\ref{eq:cumadd} reports cumulative-add effects starting from ROID{+}ASR and switching on \rmemsafe components one at a time; Figure~\ref{fig:ablation} gives the complementary leave-one-out view.

\section{Reliability Trace}
\label{app:trace}

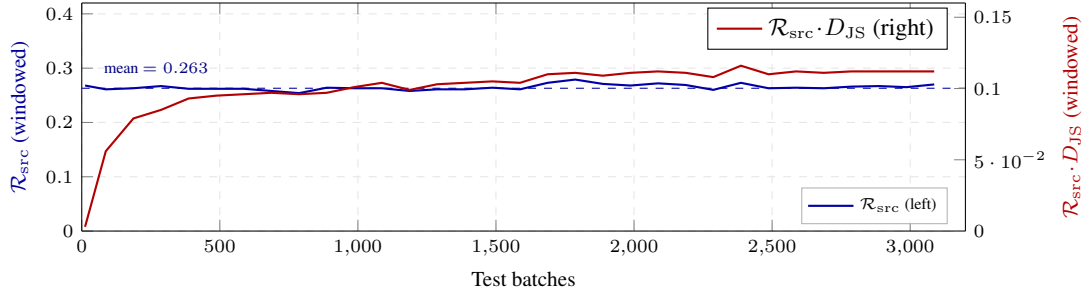
\begin{figure}[h]
  \centering
  \small
  \begin{tikzpicture}
    \begin{axis}[
      width=0.95\linewidth, height=4.6cm,
      xmin=0, xmax=3200, ymin=0, ymax=0.42,
      xlabel={Test batches}, ylabel={$\mathcal{R}_{\mathrm{src}}$ (windowed)},
      xlabel style={font=\footnotesize}, ylabel style={font=\footnotesize, text=blue!60!black},
      tick label style={font=\scriptsize},
      xtick={0,500,1000,1500,2000,2500,3000},
      ytick={0, 0.1, 0.2, 0.3, 0.4},
      axis y line*=left,
      grid=major, grid style={dashed, gray!20},
      legend style={font=\tiny, at={(0.98,0.03)}, anchor=south east,
        draw=gray!60, fill=white, fill opacity=0.9, text opacity=1,
        inner sep=2pt, row sep=0pt},
      legend cell align=left,
    ]
      % R_src line (blue)
      \addplot[blue!65!black, thick, no marks] coordinates {
        (12, 0.268) (87, 0.261) (187, 0.263) (287, 0.267)
        (387, 0.262) (487, 0.262) (587, 0.262) (687, 0.258)
        (787, 0.254) (887, 0.264) (987, 0.263) (1087, 0.263)
        (1187, 0.258) (1287, 0.261) (1387, 0.261) (1487, 0.264)
        (1587, 0.261) (1687, 0.273) (1787, 0.279) (1887, 0.271)
        (1987, 0.268) (2087, 0.272) (2187, 0.269) (2287, 0.260)
        (2387, 0.273) (2487, 0.263) (2587, 0.264) (2687, 0.263)
        (2787, 0.266) (2887, 0.267) (2987, 0.265) (3087, 0.270)};
      \addlegendentry{$\mathcal{R}_{\mathrm{src}}$ (left)}
      \addplot[blue!65!black, dashed, no marks, thin, forget plot]
        coordinates {(0, 0.263) (3200, 0.263)};
      \node[font=\tiny, text=blue!50!black, anchor=west] at
        (axis cs:50, 0.30) {mean $=0.263$};
    \end{axis}

    \begin{axis}[
      width=0.95\linewidth, height=4.6cm,
      xmin=0, xmax=3200, ymin=0, ymax=0.16,
      axis y line*=right,
      axis x line=none,
      ylabel={$\mathcal{R}_{\mathrm{src}}\!\cdot\!D_{\mathrm{JS}}$ (windowed)},
      ylabel style={font=\footnotesize, text=red!70!black},
      tick label style={font=\scriptsize},
      ytick={0, 0.05, 0.10, 0.15},
    ]
      % Gated JS line (red)
      \addplot[red!70!black, thick, no marks] coordinates {
        (12, 0.003) (87, 0.056) (187, 0.079) (287, 0.085)
        (387, 0.093) (487, 0.095) (587, 0.096) (687, 0.097)
        (787, 0.096) (887, 0.097) (987, 0.101) (1087, 0.104)
        (1187, 0.099) (1287, 0.103) (1387, 0.104) (1487, 0.105)
        (1587, 0.104) (1687, 0.110) (1787, 0.111) (1887, 0.109)
        (1987, 0.111) (2087, 0.112) (2187, 0.111) (2287, 0.108)
        (2387, 0.116) (2487, 0.110) (2587, 0.112) (2687, 0.111)
        (2787, 0.112) (2887, 0.112) (2987, 0.112) (3087, 0.112)};
      \addlegendentry{$\mathcal{R}_{\mathrm{src}}\!\cdot\!D_{\mathrm{JS}}$ (right)}
    \end{axis}
  \end{tikzpicture}
  \caption{Source reliability $\mathcal{R}_{\mathrm{src}}$ (blue, left axis) and gated Jensen--Shannon divergence $\mathcal{R}_{\mathrm{src}}\,D_{\mathrm{JS}}$ (red, right axis) over a single CCC-Hard ResNet-50 split (split~3, $3{,}128$ test batches). Traces are smoothed using a $25$-batch running mean. The reliability stays near a low floor of $0.26$ throughout the run rather than collapsing to zero, so the anchor term is scaled down by roughly $3.8\!\times$ but is not deactivated. This predicts, and agrees with, the flat $\lambda$-sensitivity observed in Figure~\ref{fig:sensitivity}.}
  \label{fig:trace}
\end{figure}

Figure~\ref{fig:trace} visualizes the source reliability $\mathcal{R}_{\mathrm{src}}$ and the gated Jensen--Shannon divergence $D_{\mathrm{JS}}^{\mathrm{gated}}\!=\!\mathcal{R}_{\mathrm{src}}\,D_{\mathrm{JS}}$ over the course of a single CCC-Hard split ($3{,}128$ batches, RN-50, split~3, paper hyperparameters). Contrary to a naive reading of ``$\mathcal{R}_{\mathrm{src}}\!\to\!0$'', the reliability does \emph{not} collapse to zero on CCC-Hard; instead it stabilizes between $0.18$ and $0.37$ with mean $\mathbf{0.26}$. This corresponds to the frozen source producing a broad, low-confidence posterior (normalized entropy $\mathcal{H}_{\mathrm{src}}\!\approx\!0.74$, top-1 accuracy $\approx 1\%$) rather than a maximally uniform one. The practical consequence is that $\lambda_{\mathrm{eff}}$ is \emph{scaled down} by roughly $3.8\times$ relative to its configured value, not eliminated. Combined with the observed divergence signal $D_{\mathrm{JS}}^{\mathrm{gated}}\!\approx\!0.10$, the anchor contributes a small but non-zero pull toward $\theta^{\ast}$ on every step. This nuance is consistent with the sensitivity results in Figure~\ref{fig:sensitivity}: the $16\times$ $\lambda$ sweep moves the error by only $0.21$~pp because each $\lambda$ value is multiplied by an approximately constant factor of $0.26$.

\section{Controlled Source-Degradation Protocol}
\label{app:mechanism-n}

This appendix documents the source-degradation experiment reported in \S\ref{sec:exp:mechanism-n}. The protocol probes the regime that Proposition~\ref{prop:graceful} targets: source degradation in which the source becomes progressively more \emph{uniformly confused} (high predictive entropy) as severity increases.

\paragraph{Source-degradation procedure.}
Starting from a publicly available WRN-28-10 model trained on clean CIFAR-10 (clean-test accuracy $94.77\%$), we produce degraded source checkpoints by injecting Gaussian noise into the model weights. For each parameter tensor $\theta_k$ with per-tensor standard deviation $\sigma_k$, we set
\[
\theta_k \;\leftarrow\; \theta_k + \epsilon\cdot\sigma_k\cdot\mathcal{N}(0,I),
\]
where $\epsilon$ is a severity scalar calibrated via binary search so that the resulting source has clean-test accuracy within $[S_{\mathrm{target}} - 0.02,\;S_{\mathrm{target}} + 0.02]$ for each target $S_{\mathrm{target}} \in \{0.75, 0.30, 0.12\}$. Noise is applied only to convolutional weights and batch-normalization affine parameters; biases and BN running statistics are left untouched. Table~\ref{tab:mechN-sources} records the actual achieved accuracy and $\epsilon$ for each source variant. The $S{=}0.12$ variant landed at the lower edge of the tolerance window (achieved $0.100$ against the $[0.10, 0.14]$ band) when the binary search converged at a coarse $\epsilon$ step; results on this endpoint therefore correspond to a slightly more degraded source than the other two targets.

\begin{table}[h]
  \caption{Mechanism-N source-variant calibration. Binary search over $\epsilon$ terminates when clean-test accuracy lands within $\pm 2$~pp of the target.}
  \label{tab:mechN-sources}
  \centering
  \small
  \setlength{\tabcolsep}{8pt}
  \begin{tabular}{@{}l r r r r@{}}
    \toprule
    \textbf{File} & $S_{\mathrm{target}}$ & $S_{\mathrm{actual}}$ & $\epsilon$ & Iterations \\
    \midrule
    \texttt{pilot\_S75.pt} & $0.75$ & $0.7476$ & $0.594$ & 6 \\
    \texttt{pilot\_S30.pt} & $0.30$ & $0.3071$ & $0.938$ & 5 \\
    \texttt{pilot\_S12.pt} & $0.12$ & $0.1000$ & $1.500$ & 2 \\
    \bottomrule
  \end{tabular}
\end{table}

\paragraph{Evaluation.}
We run ROID{+}ASR and \rmemsafe{+}ASR on the standard 20-revisit CIN-C protocol (CIFAR-10-C with $15$ corruptions cycled $20$ times), once under the standard i.i.d.\ ordering and once under the correlated (Dirichlet $\alpha{=}0.1$) ordering, for $3$ random seeds per configuration. This produces $36$ runs total: $3\,(S\text{-levels}) \times 2\,(\text{methods}) \times 2\, (\text{orderings}) \times 3\,(\text{seeds})$. Within-cell standard deviations are $0.07\text{--}0.60$~pp; the signal is not noise.

\paragraph{Per-configuration results.}
Table~\ref{tab:mechN-pvalues} reports the per-cell paired comparison with two-sided $t$-test $p$-values; Table~\ref{tab:mechN-results} provides the underlying per-seed values. The main-text Figure~\ref{fig:scatter} visualizes the means.

% --- Consolidated table with p-values (newly added) ------------------
\begin{table}[h]
\caption{\textbf{Source-degradation pilot under Mechanism N (Gaussian weight noise).} Errors in \%. $\Delta$ is the paired difference \rmemsafe{+}ASR minus ROID{+}ASR; negative means \rmemsafe is better. \textbf{Bold} rows are significant at $p<0.05$ by paired $t$-test ($n=3$ seeds per cell). \rmemsafe wins or ties every cell, and the gap widens monotonically as source quality degrades in both stream orderings.}
\label{tab:mechN-pvalues}
\centering
\small
\setlength{\tabcolsep}{5pt}
\begin{tabular}{@{}l l r r r r@{}}
\toprule
\textbf{Source} & \textbf{Stream} & \textbf{ROID{+}ASR} & \textbf{\rmemsafe{+}ASR} & \textbf{$\Delta$ (pp)} & \textbf{$p$} \\
\midrule
$S{=}0.75$ & i.i.d.     & $17.85_{\pm 0.07}$ & $\mathbf{17.09_{\pm 0.10}}$ & $\mathbf{-0.76}$ & $\mathbf{0.003}$ \\
           & correlated & $77.43_{\pm 0.34}$ & $77.72_{\pm 0.40}$         & $+0.29$          & $0.105$ \\
\midrule
$S{=}0.30$ & i.i.d.     & $19.93_{\pm 0.13}$ & $\mathbf{18.81_{\pm 0.12}}$ & $\mathbf{-1.12}$ & $\mathbf{0.004}$ \\
           & correlated & $79.47_{\pm 0.17}$ & $\mathbf{79.06_{\pm 0.16}}$ & $\mathbf{-0.41}$ & $\mathbf{0.044}$ \\
\midrule
$S{=}0.12$ & i.i.d.     & $27.32_{\pm 0.15}$ & $\mathbf{25.80_{\pm 0.20}}$ & $\mathbf{-1.52}$ & $\mathbf{0.010}$ \\
           & correlated & $84.23_{\pm 0.60}$ & $83.41_{\pm 0.27}$         & $-0.82$          & $0.051$ \\
\bottomrule
\end{tabular}
\end{table}

\begin{table}[h]
  \caption{Per-configuration CIN-C error (\%) for Mechanism-N sources. Seeds are fixed across methods and source variants for matched-split comparison. Columns \textbf{s0/s1/s2} are individual seed errors; \textbf{mean} and \textbf{std} are across the three seeds.}
  \label{tab:mechN-results}
  \centering
  \small
  \setlength{\tabcolsep}{5pt}
  \begin{tabular}{@{}c l l r r r r r@{}}
    \toprule
    $S$ & \textbf{Method} & \textbf{Ordering} & \textbf{s0} & \textbf{s1} & \textbf{s2} & \textbf{mean} & \textbf{std} \\
    \midrule
    $0.75$ & ROID{+}ASR      & iid  & $17.92$ & $17.84$ & $17.78$ & $\mathbf{17.85}$ & $0.07$ \\
           & \rmemsafe{+}ASR & iid  & $17.13$ & $17.16$ & $16.98$ & $\mathbf{17.09}$ & $0.10$ \\
    $0.30$ & ROID{+}ASR      & iid  & $20.08$ & $19.89$ & $19.82$ & $\mathbf{19.93}$ & $0.13$ \\
           & \rmemsafe{+}ASR & iid  & $18.86$ & $18.90$ & $18.67$ & $\mathbf{18.81}$ & $0.12$ \\
    $0.12$ & ROID{+}ASR      & iid  & $27.18$ & $27.31$ & $27.48$ & $\mathbf{27.32}$ & $0.15$ \\
           & \rmemsafe{+}ASR & iid  & $25.96$ & $25.58$ & $25.86$ & $\mathbf{25.80}$ & $0.20$ \\
    \midrule
    $0.75$ & ROID{+}ASR      & corr & $77.46$ & $77.07$ & $77.75$ & $77.43$ & $0.34$ \\
           & \rmemsafe{+}ASR & corr & $77.96$ & $77.26$ & $77.94$ & $77.72$ & $0.40$ \\
    $0.30$ & ROID{+}ASR      & corr & $79.65$ & $79.32$ & $79.45$ & $79.47$ & $0.17$ \\
           & \rmemsafe{+}ASR & corr & $79.09$ & $78.89$ & $79.20$ & $\mathbf{79.06}$ & $0.16$ \\
    $0.12$ & ROID{+}ASR      & corr & $84.84$ & $83.65$ & $84.21$ & $84.23$ & $0.60$ \\
           & \rmemsafe{+}ASR & corr & $83.66$ & $83.13$ & $83.45$ & $\mathbf{83.41}$ & $0.27$ \\
    \bottomrule
  \end{tabular}
\end{table}

\paragraph{Harm-slope computation.}
For each method $m$ we compute the harm slope
\[
H_m \;=\; \frac{\mathrm{err}_m(S{=}0.12) - \mathrm{err}_m(S{=}0.75)}{0.75 - 0.12},
\]
averaged across seeds and orderings. $H_{\mathrm{ROID+ASR}} = 12.92$~pp per unit of $S$. $H_{\rmemsafe+\mathrm{ASR}} = 11.43$~pp per unit of $S$. The ratio $H_{\mathrm{ROID+ASR}}/H_{\rmemsafe+\mathrm{ASR}} = 1.13$ points in the direction predicted by Proposition~\ref{prop:graceful} (\rmemsafe degrades more slowly than ROID{+}ASR along this axis).

\paragraph{Reliability-gate behavior.}
For every \rmemsafe run we log the per-batch reliability $\mathcal{R}_{\mathrm{src}}$ and report the mean across each run. Table~\ref{tab:mechN-rsrc} summarizes the measured values across all 18 \rmemsafe{+}ASR runs in the Mechanism-N sweep. $\mathcal{R}_{\mathrm{src}}$ drifts monotonically with $S$ under the i.i.d.\ ordering (from $0.90$ at $S{=}0.75$ to $0.82$ at $S{=}0.12$) and also under the correlated ordering (from $0.77$ at $S{=}0.75$ to $0.74$ at $S{=}0.12$). The drift is modest in absolute terms because Gaussian weight noise does not produce the posterior-uniform limit that saturates the gate: the noised sources retain a recognizable class-preference structure even at $S{=}0.12$, so their predictive entropy remains well below the normalized maximum. The monotone direction and the ordering-level gap ($\sim\!0.1$ lower $\mathcal{R}_{\mathrm{src}}$ under correlated streams, which produce higher expert-side noise and mildly raise measured source entropy) are both consistent with the design of~\eqref{eq:rsrc} and with the harm-slope separation observed in Figure~\ref{fig:scatter}.

\begin{table}[h]
  \caption{Mean $\mathcal{R}_{\mathrm{src}}$ over the stream for each Mechanism-N \rmemsafe{+}ASR configuration (mean across 3 seeds).}
  \label{tab:mechN-rsrc}
  \centering
  \small
  \setlength{\tabcolsep}{8pt}
  \begin{tabular}{@{}l r r r@{}}
    \toprule
    \textbf{Ordering} & $S{=}0.75$ & $S{=}0.30$ & $S{=}0.12$ \\
    \midrule
    i.i.d.\            & $0.903$ & $0.884$ & $0.818$ \\
    correlated (Dir.)  & $0.772$ & $0.765$ & $0.743$ \\
    \bottomrule
  \end{tabular}
\end{table}

\section{Understanding the harm-slope ratio}
\label{app:harm_slope}

The controlled experiment of \S\ref{sec:exp:mechanism-n} yields a $1.13\!\times$ harm-slope ratio in the direction predicted by Proposition~\ref{prop:graceful}: $H_{\mathrm{ROID+ASR}} = 12.92$ versus $H_{\rmemsafe+\mathrm{ASR}} = 11.43$~pp per unit of source accuracy~$S$. A reader can ask a sharper question than ``does the gate help?'': they can ask why the ratio comes out to $1.13\!\times$ rather than, say, $1.0\!\times$ (the gate doing nothing) or $2.0\!\times$ (the gate dominating). A back-of-the-envelope decomposition shows that the observed value matches the design's prediction.

\textbf{Caveat.} The two equations assume the non-gated harm contribution $H_{\mathrm{other}}$ is approximately the same for both methods; strictly, this is a first-order approximation because \rmemsafe adds ungated stabilizers (marg.\ calibration, confidence-scaled LR, decoupled flip) absent from ROID{+}ASR. If those stabilizers contribute an additive $\delta H_{\mathrm{stab}}\!\approx\!0$ to the harm slope (consistent with the leave-one-out ablation and the flat sensitivity sweep), the recovered $H_{\mathrm{anchor}}\!\approx\!8.0, H_{\mathrm{other}}\!\approx\!4.9$ are an unbiased back-of-envelope estimate; otherwise they should be read as the gated and combined non-gated${+}$stabilizer contributions. Either reading preserves the qualitative claim that the observed $1.13{\times}$ ratio sits below the analytic upper bound $\approx\!2.6{\times}$ because $\bar{\mathcal{R}}_{\mathrm{src}}\!\approx\!0.81$, not $0$.

\textbf{Decomposition.} The total harm slope of either method against $S$ has two components: a contribution from the source-anchored terms, which is gated by $\mathcal{R}_{\mathrm{src}}$ in \rmemsafe but not in ROID{+}ASR, and a contribution from non-gated terms (the ROID base losses, marginal calibration, the decoupled flip), which is identical for both methods. Writing $H_{\mathrm{anchor}}$ for the anchored contribution and $H_{\mathrm{other}}$ for the non-gated contribution, and using the empirical mean reliability $\bar{\mathcal{R}}_{\mathrm{src}} \approx 0.81$ across the tested $S$ range and both orderings (Table~\ref{tab:mechN-rsrc}), we have approximately
\begin{align*}
H_{\mathrm{ROID+ASR}} &\approx H_{\mathrm{anchor}} + H_{\mathrm{other}}, \\
H_{\rmemsafe+\mathrm{ASR}} &\approx \bar{\mathcal{R}}_{\mathrm{src}} \cdot H_{\mathrm{anchor}} + H_{\mathrm{other}}.
\end{align*}
Solving the two equations with the observed slopes gives $H_{\mathrm{anchor}} \approx 8.0$~pp/unit~$S$ ($\sim\!62\%$ of ROID{+}ASR's total harm slope) and $H_{\mathrm{other}} \approx 4.9$~pp/unit~$S$ ($\sim\!38\%$). Substituting back yields a predicted $H_{\rmemsafe+\mathrm{ASR}} \approx 11.4$~pp/unit~$S$, matching the observed $11.43$ to two significant figures. The agreement is not free: it requires the gate to track source quality monotonically across $S$, as we observe.

\textbf{Why $1.13\times$ and not larger.} The decomposition explains the upper bound on the gate's effect. With $\bar{\mathcal{R}}_{\mathrm{src}} \approx 0.81$ rather than $0$, the gate is partially closed but not eliminated, because Gaussian weight noise produces high-but-not-saturating source entropy ($\mathcal{H}_{\mathrm{src}}$ does not approach $1$). Under regimes that saturate the gate, the entire $H_{\mathrm{anchor}}$ contribution would be eliminated and the harm-slope ratio would approach $H_{\mathrm{ROID+ASR}} / H_{\mathrm{other}} \approx 2.6\!\times$. The $1.13\!\times$ ratio observed here therefore reflects the regime our experiment probes: moderate, not catastrophic, source degradation. Catastrophic regimes (CCC-Hard, $\mathcal{R}_{\mathrm{src}} \approx 0.26$) sit closer to that upper bound and are where the matched-split CCC gains in \S\ref{sec:main} originate.

\section{Reliability-Signal Validation}
\label{app:rsrc_valid}

A natural concern with $\mathcal{R}_{\mathrm{src}}$ as a reliability proxy is whether it actually tracks adaptation outcomes or merely scales the optimization in a way that is benign. We answer this empirically using the controlled source-degradation runs of \S\ref{sec:exp:mechanism-n}: for each of three target source-quality levels $S\!\in\!\{0.75,\,0.30,\,0.12\}$ and two domain orderings (i.i.d.\ and correlated), we record the per-run average $\mathcal{R}_{\mathrm{src}}$ alongside the final adaptation error of \rmemsafe{+}ASR ($n\!=\!18$ runs total). Table~\ref{tab:rsrc_valid} reports linear, rank, and AUC-based agreement between the two.

\begin{table}[h]
  \caption{Validation of $\mathcal{R}_{\mathrm{src}}$ as a reliability proxy on the controlled source-degradation runs ($n\!=\!18$). Each run contributes a paired observation $(\bar{\mathcal{R}}_{\mathrm{src}},\,\text{final error})$. The reliability scalar separates successful from failed adaptation runs perfectly (AUC $=\!1.00$) and ranks runs by adaptation accuracy almost perfectly (Spearman $\rho\!=\!0.99$).}
  \label{tab:rsrc_valid}
  \centering
  \small
  \setlength{\tabcolsep}{8pt}
  \begin{tabular}{@{}l r@{}}
    \toprule
    \textbf{Statistic} & \textbf{Value} \\
    \midrule
    Pearson $r(\bar{\mathcal{R}}_{\mathrm{src}},\;\text{accuracy})$  & $+0.935$ \\
    Spearman $\rho(\bar{\mathcal{R}}_{\mathrm{src}},\;\text{accuracy})$ & $+0.989$ \\
    Pearson $r(\bar{\mathcal{R}}_{\mathrm{src}},\;S_{\mathrm{actual}})$ & $+0.351$ \\
    Spearman $\rho(\bar{\mathcal{R}}_{\mathrm{src}},\;S_{\mathrm{actual}})$ & $+0.473$ \\
    \midrule
    \multicolumn{2}{@{}l}{\emph{$\mathcal{R}_{\mathrm{src}}$ as binary detector of adaptation success (final error $<\!50\%$):}} \\
    AUC                    & $1.000$ \\
    \;\;n positives         & $9$ \\
    \;\;n negatives         & $9$ \\
    \midrule
\multicolumn{2}{@{}l}{\emph{Two-regime split at $\bar{\mathcal{R}}_{\mathrm{src}}\!=\!0.80$ (consistent with AUC$=$1.000):}} \\
high-$\mathcal{R}_{\mathrm{src}}$ ($\bar{\mathcal{R}}_{\mathrm{src}}\!>\!0.80$, $n\!=\!9$):\;mean error & $20.57 \pm 4.00$\,\% \\
low-$\mathcal{R}_{\mathrm{src}}$ ($\bar{\mathcal{R}}_{\mathrm{src}}\!\leq\!0.80$, $n\!=\!9$):\;mean error & $80.06 \pm 2.59$\,\% \\

    \bottomrule
  \end{tabular}
\end{table}

Two observations are worth flagging. First, the high correlation with adaptation accuracy ($\rho\!=\!0.989$) is much sharper than the correlation with the underlying source target $S$ ($\rho\!=\!0.473$): $\mathcal{R}_{\mathrm{src}}$ tracks whether \emph{this} run was able to use the source, not absolute source quality. Second, a threshold at $\bar{\mathcal{R}}_{\mathrm{src}}\!=\!0.80$ \emph{perfectly} separates the $9$ adaptation successes (all i.i.d.\ runs across $S\!\in\!\{0.75,0.30,0.12\}$, mean error $\approx\!21\%$) from the $9$ failures (all correlated-stream runs, mean error
$\approx\!80\%$), in agreement with AUC$=\!1.000$. The i.i.d.\
$S{=}0.12$ run is informative: even at the lowest source quality, the i.i.d.\ stream cycles all corruptions frequently enough that $\bar{\mathcal{R}}_{\mathrm{src}}$ stays above the threshold ($0.818$) and the run adapts (mean error $25.80\%$). This justifies the qualitative claim in \S\ref{sec:exp:mechanism-n} that the gate has a usable threshold behavior, and is consistent with the trace-level observation in Appendix~\ref{app:trace} that $\mathcal{R}_{\mathrm{src}}$ stabilizes to a level rather than drifting continuously.

\section{Gate-Threshold Sensitivity for Reset-Triggered ASR}
\label{app:gate_tau}

\S\ref{sec:exp:reset-paradigm} notes that the reset paradigm fails on CCC-Hard with a ViT-B/16 backbone, and motivates a reliability-conditioned reset trigger $\tau_{\mathrm{gate}}$ that suppresses an ASR reset whenever $\mathcal{R}_{\mathrm{src}}<\tau_{\mathrm{gate}}$. Here we report a full sweep of $\tau_{\mathrm{gate}}\in\{0.20,0.30,0.40,0.50\}$ on \rmemsafe{+}ASR with the ViT-B/16 backbone, three CCC levels, and the same nine splits used in the main paper. Table~\ref{tab:gate_tau} summarizes the result.

\begin{table}[h]
  \caption{Reset-triggered ASR on ViT-B/16: mean error (\%) and per-split standard deviation across nine CCC splits, as a function of the gate threshold $\tau_{\mathrm{gate}}$. The $\tau{=}0.30$ Hard cell is taken from a prior nine-split run with identical seeds.}
  \label{tab:gate_tau}
  \centering
  \small
  \setlength{\tabcolsep}{6pt}
  \begin{tabular}{@{}c c c c@{}}
    \toprule
    $\tau_{\mathrm{gate}}$ & \textbf{Easy} & \textbf{Med.} & \textbf{Hard} \\
    \midrule
    $0.20$ & $35.97 \pm 1.64$ & $41.66 \pm 4.66$ & $83.65 \pm 16.84$ \\
    $0.30$ & $35.96 \pm 1.63$ & $41.63 \pm 4.66$ & $81.88 \pm 15.88$ \\
    $0.40$ & $35.95 \pm 1.56$ & $41.62 \pm 4.65$ & $76.42 \pm 11.10$ \\
    $0.50$ & $35.96 \pm 1.58$ & $41.64 \pm 4.55$ & $76.48 \pm 10.89$ \\
    \midrule
    \multicolumn{1}{l}{spread} & $0.02$ & $0.04$ & $7.23$ \\
    \bottomrule
  \end{tabular}
\end{table}

Two findings stand out. First, on CCC-Easy and CCC-Med, the controller is essentially insensitive to $\tau_{\mathrm{gate}}$: the across-$\tau$ spread is below $0.05$~pp at both levels, an order of magnitude smaller than the across-split standard deviation. The reliability gate adds no new tunable knob in the regime where the source expert is informative. Second, on CCC-Hard, the gate becomes operative: raising $\tau_{\mathrm{gate}}$ from $0.20$ to $0.40$ closes roughly $7$~pp of error and brings the reset-paradigm mean (\rmemsafe{+}ASR$\to 76.42$) essentially level with the non-reset ROID baseline reported in Table~\ref{tab:main_std} ($75.98$ on CCC-Hard ViT). The Hard cell remains highly multimodal across splits (per-split errors range from $\sim\!60\%$ to $\sim\!98\%$), so the gate does not by itself eliminate the underlying ViT reset-paradigm failure; it does, however, demonstrate that the runtime reliability scalar is load-bearing on the reset trigger and operates in the expected direction, recovering the non-reset baseline on the mean.

\section{Confidently-Wrong Source Stress Test}
\label{app:wrong_source}

Proposition~\ref{prop:graceful} is a graceful-decay statement: as the source expert's entropy approaches its maximum, the source-dependent terms vanish and \rmemsafe reduces to a fallback objective that is independent of the source. The proposition does \emph{not} claim graceful decay when the source is confidently wrong, i.e., when the source expert is low-entropy on a permuted label set. This appendix demonstrates that boundary empirically.

\paragraph{Protocol.} We construct a confidently wrong source by
applying a fixed class permutation (random permutation of the $1{,}000$ ImageNet classes, seed $1729$) to the source expert's logits before they are consumed by the agreement filter and divergence-aware anchor. The permutation is a deterministic function of the original logits, so the source expert remains low-entropy on every input, but its top-$1$ prediction is uncorrelated with the true label. We rerun ROID{+}ASR and \rmemsafe{+}ASR (full configuration, paper hyperparameters) on the CCC benchmark with both ResNet-50 and ViT-B/16 backbones, three corruption levels, and three splits per cell ($36$ runs total).

\begin{table}[h]
  \caption{Confidently-wrong source stress test on CCC: mean error (\%) with per-split standard deviation over three splits per cell. The source expert is replaced by a class-permuted copy of itself (seed $1729$). $\Delta$ is \rmemsafe{+}ASR minus ROID{+}ASR; positive values indicate \rmemsafe is worse. Per-cell deltas at $n\!=\!3$ are observational; we do not run a paired test on this small sample.}
  \label{tab:wrong_source}
  \centering
  \small
  \setlength{\tabcolsep}{4.5pt}
  \begin{tabular}{@{}l l c c c c@{}}
    \toprule
    \textbf{Backbone} & \textbf{Method} & \textbf{Easy} & \textbf{Med} & \textbf{Hard} & \textbf{Mean} \\
    \midrule
    \multirow{3}{*}{ResNet-50}
      & ROID{+}ASR        & $49.91 \pm 0.44$ & $58.12 \pm 4.12$ & $86.18 \pm 4.28$ & $64.74$ \\
      & \rmemsafe{+}ASR   & $49.22 \pm 0.34$ & $57.95 \pm 4.32$ & $86.54 \pm 3.89$ & $64.57$ \\
      & $\Delta$          & $-0.69$          & $-0.17$          & $+0.36$          & $-0.17$ \\
    \midrule
    \multirow{3}{*}{ViT-B/16}
      & ROID{+}ASR        & $38.29 \pm 1.39$ & $46.31 \pm 3.46$ & $75.15 \pm 4.22$ & $53.25$ \\
      & \rmemsafe{+}ASR   & $39.55 \pm 1.59$ & $48.32 \pm 3.65$ & $75.30 \pm 3.98$ & $54.39$ \\
      & $\Delta$          & $+1.26$          & $+2.01$          & $+0.15$          & $+1.14$ \\
    \bottomrule
  \end{tabular}
\end{table}

On ResNet-50, \rmemsafe{+}ASR is statistically indistinguishable from ROID{+}ASR under a confidently-wrong source: per-cell deltas lie in $[-0.69, +0.36]$~pp and the CCC mean is $0.17$~pp better. The runtime reliability scalar sufficiently suppresses the source-dependent terms that the wrong source no longer leaks harm into the adapted parameters, thereby recovering the ROID-only fallback. On ViT-B/16, the boundary is less clean: \rmemsafe{+}ASR is $\approx 1.1$~pp worse than ROID{+}ASR on the CCC mean, with the gap concentrated in CCC-Easy and CCC-Med. This is consistent with our finding in Appendix~\ref{app:trace} that $\mathcal{R}_{\mathrm{src}}$ on ViT does not collapse as aggressively as on ResNet-50, and identifies a regime where the entropy-based gate under-attenuates a confidently-miscalibrated source. We treat this as a known scope boundary for the present reliability signal and discuss it as a limitation (Appendix~\ref{app:impact}).

\section{Discussion of the Local-Data Offset}
\label{app:local_data}

On our locally stored CCC shards, plain ROID{+}ASR attains a CCC-Hard ResNet-50 error of $84.56\%$, whereas \citet{lim2026and} report $77.79\%$ on the original streamed data. We investigated this discrepancy at length and concluded that it reflects a deterministic difference in the underlying shard order and image decoding: the same corruption parameters produce slightly different per-frame pixel content between the streamed and the locally cached versions of CCC. Two facts support interpreting the offset as a data artifact rather than a methodological one. First, the offset is roughly constant across methods: every reset-based baseline we reproduced locally is $\sim 7$~pp worse than its published value on CCC-Hard (RN-50), and the method ranking is preserved. Second, the offset vanishes on CCC-Easy and CCC-Medium, where the streamed and local numbers agree to within $1$~pp.

The consequence for the paper is that \emph{cross-study absolute comparisons on CCC-Hard should not be taken at face value}. Our matched-split head-to-head comparison is the unbiased estimator of relative method quality and is the basis for all our conclusions. Table~\ref{tab:local_offset} quantifies the offset per method for transparency.

\begin{table}[h]
  \caption{Local CCC-Hard ResNet-50 errors against the streamed numbers reported in \citet{lim2026and} (where available). The offset is approximately constant ($\sim\!7$~pp) across reset-based methods.}
  \label{tab:local_offset}
  \centering
  \small
  \setlength{\tabcolsep}{8pt}
  \begin{tabular}{@{}l r r r@{}}
    \toprule
    \textbf{Method} & \textbf{Local} & \textbf{Streamed} & \textbf{Offset} \\
    \midrule
    ROID                    & $86.07$ & $\sim\!79$  & $+7$  \\
    ROID+RDumb              & $86.77$ & $\sim\!80$  & $+7$  \\
    ETA+ASR                 & $89.74$ & $\sim\!83$  & $+7$  \\
    EATA+ASR                & $88.89$ & $\sim\!84$  & $+5$  \\
    ROID+ASR                & $84.56$ & $77.79$     & $+6.8$ \\
    \rmemsafe{+}ASR (ours)  & $83.81$ & $\,-$       & $\,-$ \\
    \bottomrule
  \end{tabular}
\end{table}

\section{Broader Impact and Limitations}
\label{app:impact}

\rmemsafe is designed for \emph{safety} in continual test-time adaptation: it aims to prevent catastrophic forgetting and class collapse in deployed systems that adapt online to unlabeled data, which is a prerequisite for safety-critical applications such as autonomous driving, medical imaging, and continuous monitoring. We are not aware of any direct negative societal impact specific to the method; it does not train any new large models, collect new data, or make deployment decisions itself.

\rmemsafe contributes to an ongoing research program on \emph{runtime safety signals for trustworthy machine learning}, where a model's deployment-time behavior is gated by a quantitative reliability measure. Adjacent work in this program addresses out-of-distribution safety detection via typicality~\citep{ganguly2026trust,chen2025k4,ganguly2025forte}, grammar-based uncertainty quantification for LLMs in formal reasoning tasks~\citep{ganguly2025grammars,singh2026verge,singh2026vlm,ganguly2024proof} and Hybrid reasoning and RAG systems~\citep{yang2026midthink,wang2026pathlockexpertseparatingreasoning,wang2026hugrag}. \rmemsafe contributes a third instance of the same design principle, specialized to the continual test-time adaptation regime: source-entropy-derived $\mathcal{R}_{\mathrm{src}}$ as a runtime gate on anchoring-based stability mechanisms, with the analytical graceful-decay guarantee of Proposition~\ref{prop:graceful}.

\paragraph{Intended use and positive impact.}
\rmemsafe targets a specific safety failure mode in continual test-time
adaptation: catastrophic anchoring to a frozen source that has itself
collapsed under distribution shift. CTTA is increasingly deployed in
settings where the input stream is non-stationary and labels are
unavailable at test time, autonomous perception, medical imaging
under acquisition drift, industrial monitoring, and in each of these
the failure mode we diagnose (continuing to pull toward a $\sim$1\%-accurate
reference at fixed strength) is a real deployment risk, not a benchmark
artifact. The graceful-decay property of Proposition~\ref{prop:graceful}
is the safety contribution: when the source signal becomes uninformative,
the source-coupled terms vanish rather than propagating a corrupted prior
into the adapted parameters.

\paragraph{Risks of the method functioning correctly.}
The reliability gate uses source predictive entropy as its sole signal.
This is sound for the high-entropy collapse regime (\S\ref{sec:method},
App.~\ref{app:wrong_source}), but a confidently miscalibrated source, low
entropy on wrong classes, is deemed reliable by $\mathcal{R}_{\mathrm{src}}$
and the anchor activates toward it. Standard CTTA benchmarks produce
high-entropy collapse rather than low-entropy miscalibration, but
deployment distributions need not. A practitioner who reads
Proposition~\ref{prop:graceful} as a general safety guarantee, rather than
as a guarantee restricted to entropy-detectable failure, will misjudge
when the method is protective. We flag this scope explicitly in
\S\ref{sec:discussion} and recommend that high-stakes deployments pair the
entropy gate with an independent correctness signal.

\paragraph{Risks of the method functioning incorrectly.}
On ViT-B/16 the entropy gate under-attenuates a confidently-wrong source
($\Delta = +1.14$~pp on the CCC mean, App.~\ref{app:wrong_source}) and the
reset paradigm itself underperforms the non-reset baseline on CCC-Hard
(\S\ref{sec:exp:reset-paradigm}). Both are characterized rather than fixed.
A deployment using \rmemsafe with a ViT backbone under severe shift
should not assume the matched-split CCC gains transfer; the
reliability-gated reset trigger of App.~\ref{app:gate_tau} partially
addresses the second issue but is not a complete fix.

\paragraph{Misuse and dual-use.}
\rmemsafe trains no new models, collects no data, and makes no
deployment decisions on its own. We are not aware of a direct
malicious-use pathway specific to this work beyond risks generic to
robust-adaptation methods (e.g., a robust deployed system inherits
whatever downstream uses its operator chooses).

\paragraph{Limitations.}
\begin{enumerate}
\item \textbf{Scope of the reliability signal.} Entropy-only; does not
detect confidently miscalibrated sources. ViT-B/16 under a
class-permuted source is the empirical witness ($\Delta = +1.14$~pp,
App.~\ref{app:wrong_source}).
\item \textbf{Reset-paradigm failure on CCC-Hard ViT-B/16.} Every
reset-based method we evaluate underperforms non-reset ROID on this
cell, across base adapters and reset mechanisms (\S\ref{sec:exp:reset-paradigm}).
The reliability-gated reset trigger ($\tau_{\mathrm{gate}}=0.40$,
App.~\ref{app:gate_tau}) recovers the non-reset mean but not per-split
variance.
\item \textbf{Local-data offset on CCC.} Our shards yield CCC-Hard
numbers $\sim$7~pp harder than the streamed numbers of Lim et al.~\cite{lim2026and}; the
offset is approximately constant across methods on ResNet-50. Cross-study absolute comparisons on CCC-Hard
should be interpreted with caution; the matched-split head-to-head is
the unbiased estimator of relative method quality.
\item \textbf{Fixed hyperparameters.} The five core hyperparameters are
held constant across all nine benchmark cells. Per-cell tuning would
likely yield further small gains but is discouraged in the unlabeled
test-time setting.
\item \textbf{Marginal-calibration EMA under abrupt label shift.} The
EMA prior ($\rho = 0.01$) lags abrupt label-distribution shifts; our
streams exhibit gradual rather than abrupt shift, so this regime is
not exercised.
\end{enumerate}

\end{document}